\PassOptionsToPackage{table}{xcolor}
\documentclass{article} 
\usepackage[T1]{fontenc}
\usepackage{iclr2027_conference,times}

\usepackage{amsmath,amsfonts,bm}

\def\eqref#1{equation~\ref{#1}}

\def\1{\bm{1}}

\DeclareMathAlphabet{\mathsfit}{\encodingdefault}{\sfdefault}{m}{sl}
\SetMathAlphabet{\mathsfit}{bold}{\encodingdefault}{\sfdefault}{bx}{n}

\usepackage{algorithm}
\usepackage{algpseudocode}
\usepackage[colorlinks=true,citecolor=blue!55!black,urlcolor=black]{hyperref}
\usepackage{url}
\usepackage{graphicx}
\usepackage{booktabs}
\usepackage{multirow}
\usepackage{tabularx}
\makeatletter
\let\roadmopd@endtabular\endtabular
\usepackage{arydshln}
\let\endtabular\roadmopd@endtabular
\expandafter\let\csname endtabular*\endcsname\endtabular
\makeatother
\usepackage{wrapfig}
\usepackage{afterpage}
\usepackage{placeins}
\usepackage{capt-of}
\usepackage{xspace}

\newcommand{\method}{\mbox{\textsc{Road MOPD}}\xspace}

\title{Beyond Domain-Routed Supervision:\\
Cross-Teacher Guarding for Multi-Teacher\\
On-Policy Distillation}

\author{%
\textbf{Shi-Yu~Tian$^{1,2,*}$, Yong~Liu$^{2,*}$, Ximan~Liu$^{2,\dagger}$, Tianrui~Zhu$^{2}$, Sule~Bai$^{2}$,} \\
\textbf{Xinwei~Yang$^{2}$, Ruiqi~Liu$^{2}$, Kun-Yang~Yu$^{1}$, Ming~Yang$^{1}$,} \\
\textbf{Zhi~Zhou$^{1,\ddagger}$, Yu-Feng~Li$^{1,\ddagger}$, Yipeng~Sun$^{2}$, Simiu~Gu$^{2}$,} \\[4pt]
\normalfont $^{1}$Nanjing University\quad $^{2}$JD.com \\[3pt]
\normalfont\small $^{*}$Equal contribution\quad $^{\dagger}$Project leader\quad $^{\ddagger}$Corresponding authors
}

\iclrfinalcopy

\begin{document}

\maketitle
\lhead{Paper Under Review}

\begin{abstract}
Multi-teacher on-policy distillation (MOPD) extends on-policy distillation (OPD) by routing domain-specific data to specialized teachers, integrating their capabilities through dense token-level supervision on student-generated trajectories.
However, domain
routing only distinguishes the supervision sources for different domains;
updates from all domains still act on shared parameters. When teachers differ
substantially in response style and domain expertise, supervision that is
effective in isolation may still lead to persistent gradient conflicts and
capability degradation during joint optimization. To address this challenge,
we propose C\textbf{ro}ss-Teacher Gu\textbf{a}r\textbf{d}ing for MOPD
(\method), which introduces non-routed teachers as
auxiliary cross-domain critics on the same student trajectories and coordinates
multi-teacher supervision through token-level credit assignment and gradient
updates. Specifically, cross-teacher-guided credit assignment uses teacher
agreement to strengthen token-level corrections jointly supported by the
teachers and attenuate conflicting signals; preventive multi-teacher gradient
projection uses auxiliary gradients from the same trajectories to constrain
potentially conflicting updates and adaptively controls projection strength
to retain target-domain learning signals. In experiments, \method
mitigates degradation in the disadvantaged domain, gains over
5 percentage points under high conflict, and improves overall performance
in low-conflict and three-domain settings.


\end{abstract}

\section{Introduction}

Reinforcement learning (RL) has emerged as an important paradigm for post-training
large language models~\citep{GPT4,guo2025deepseekr1}. However, training with
outcome-level rewards lacks direct feedback on intermediate reasoning steps,
making it difficult to attribute a response's success or failure to specific
generation decisions. Optimization therefore remains largely dependent on policy
exploration, posing challenges for credit assignment and stable optimization.
On-policy distillation (OPD)~\citep{minillm2024,agarwal2024OPD,distillm2025}
offers a promising alternative for addressing this limitation: it leverages a
strong teacher to provide dense token-level supervision on trajectories generated
by the student's current policy, combining on-policy sampling aligned with the
student's current generation distribution with fine-grained teacher feedback.

Prior work has improved OPD capability transfer through better distillation
objectives and optimization
strategies~\citep{li2026filter,lin2026opdwithverifiablereward,
yang2026filteropd,zhang2026teacherlikelihoodgroupcalibratedonpolicy}. A natural
extension is to consolidate specialized teachers' complementary capabilities
in a single student across domains.
Multi-teacher on-policy distillation (MOPD)~\citep{ma2026mopdmultiteacheronpolicydistillation,xu2026deepseekv4} pursues this goal, typically through
direct domain routing: a sample from domain $d$ receives supervision only from its
routed teacher $T_d$, while updates from all domains act on the same student
model. This design exploits each teacher's domain expertise but does not
explicitly coordinate updates from different domains to shared parameters.
Consequently, supervision that is effective in single-domain training may
induce interfering update directions during joint optimization, limiting
multi-domain capability integration.

\afterpage{%
\begin{figure}[t]
    \centering
    \includegraphics[width=\textwidth]{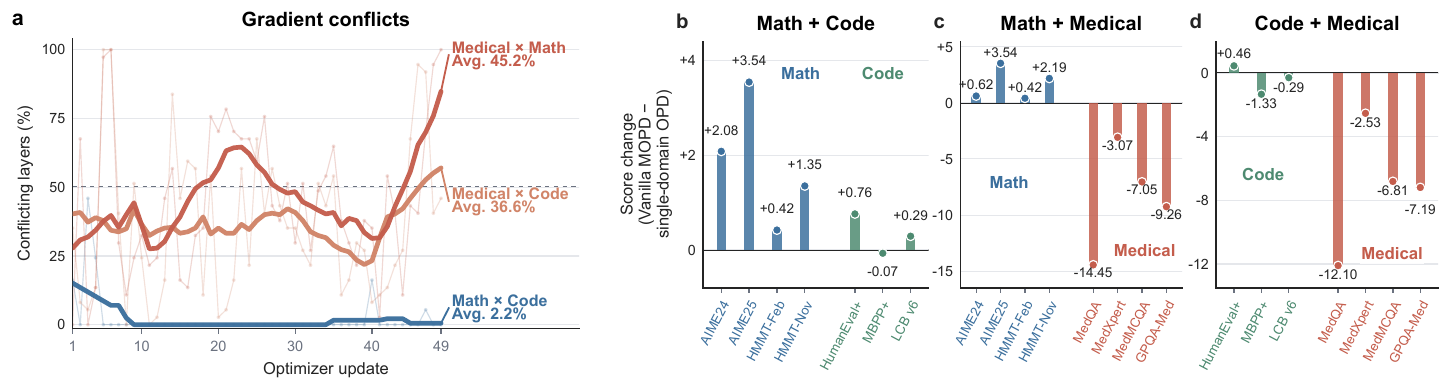}
    \caption{\textbf{Update conflicts reveal a failure mode of domain-routed MOPD.}
    (a) Fraction of conflicting layers during training; labels show training averages.
    (b--d) Per-benchmark score changes of Vanilla MOPD relative to single-domain OPD.
    Math--Code remains compatible, whereas both Medical pairings consistently
    degrade Medical performance.}
    \label{fig:gradient-conflict}
\end{figure}%
}

Prior MOPD studies have demonstrated the potential of joint learning on domain
combinations such as math and code~\citep{yang2026gopd,hou2026uniopd}, but these gains may not generalize to
other combinations. As shown in Figure~\ref{fig:gradient-conflict}, joint training
on math and code generally preserves or improves upon the corresponding
single-domain OPD performance, with relatively few gradient conflicts. In
contrast, jointly training medicine with either math or code consistently
yields a high proportion of layers with conflicting gradients, and this proportion
increases further in later stages of training. In both pairings, medical
performance falls below the single-domain OPD baseline, showing that the
joint-learning benefits observed for math and code do not extend to these
combinations. This contrast highlights
compatibility between domain updates as an important factor in MOPD and suggests
that allocating supervision through domain routing alone is insufficient to
coordinate capability transfer through shared parameters.

To address this limitation, we reconsider teachers' roles in MOPD: can
non-routed teachers help anticipate and mitigate update interference while
preserving the routed teacher's domain-specific supervision? We retain
this supervision and introduce non-routed teachers as cross-domain critics
on the same student-generated trajectories. Their feedback on the same
behavior provides local cues about potentially incompatible learning signals.
We aim to reduce joint-optimization interference through this feedback while
retaining target-domain learning signals.


Specifically, we propose C\textbf{ro}ss-Teacher Gu\textbf{a}r\textbf{d}ing for MOPD (\method),
to exploit cross-teacher feedback at both the token and
gradient levels. At the token level, \method compares evidence from the routed
and auxiliary teachers relative to a shared frozen reference model, strengthening
supervision when their implied correction directions agree and
downweighting the corresponding tokens when they conflict. At the gradient level,
auxiliary supervision on the same trajectory defines a cross-domain reference
direction, and preventive gradient projection suppresses components of the
target-domain update that conflict with this direction. In high-conflict
settings, \method improves Medical accuracy by 5.6--8.8 percentage points
over Vanilla MOPD. It also improves overall performance in low-conflict
and three-domain settings.

Our contributions are as follows:
\begin{itemize}
    \item We identify a failure mode of domain-routed MOPD: when domain updates
    are poorly compatible, separately assigned supervision can induce persistent
    gradient conflicts in shared parameters. These conflicts accompany asymmetric
    performance degradation, highlighting the importance of update compatibility
    for joint learning.

    \item We propose \method, using non-routed teachers as cross-domain critics
    on the same student trajectories. Cross-teacher agreement guides token
    weighting, while preventive gradient projection adapts its strength under
    a norm-retention constraint to protect target-domain learning signals.

    \item Across domain combinations and teacher--student configurations,
    \method improves adversely affected domains under high conflict, preserves
    low-conflict gains, and generalizes to three domains, outperforming standard
    MOPD and gradient-conflict correction baselines in overall performance.
\end{itemize}




\section{Related Work}

\subsection{On-Policy Distillation}

Classical knowledge distillation trains a student to match a teacher's predictive
distribution \citep{hinton2015KD}. Sequence-level distillation applies this idea to
autoregressive models using teacher-generated responses, but creates a mismatch
between training and inference prefixes \citep{kim2016eqkd}. On-policy distillation
(OPD) instead queries the teacher on student-generated trajectories, using
reverse-KL, general on-policy, or contrastive objectives
\citep{minillm2024,agarwal2024OPD,distillm2025}. Subsequent work analyzes
sampled-token estimators \citep{shao2026tokenlevelanalys}; makes supervision more
selective through token filtering, reweighting, verifiable rewards, or reasoning
progress \citep{li2026filter,lin2026opdwithverifiablereward,yang2026filteropd,
zhang2026teacherlikelihoodgroupcalibratedonpolicy}; and extends OPD across
tokenizers or intermediate generation steps
\citep{he2026simpleopdsimpletokenizeragnosticonpolicy,
sun2026steplevelonpolicydistillationinterpolating}. Reference-based extrapolation
further scales teacher corrections against a frozen student initialization
\citep{yang2026gopd}. Despite these advances, OPD can generalize unevenly across
capabilities \citep{li2026coinsidesdualnature}. These methods refine how evidence
from one assigned teacher is estimated, selected, or scaled; they do not test its
compatibility with capabilities represented by other teachers. We instead study
how domain-specialized teachers can jointly assess the same student-generated
prefixes.

\subsection{Multi-Teacher On-Policy Distillation}

Multi-teacher on-policy distillation (MOPD) transfers capabilities from domain
specialists to a shared student, typically routing each rollout to its
corresponding teacher for dense supervision
\citep{ma2026mopdmultiteacheronpolicydistillation}. Domain routing is also used in
large-scale and staged multi-domain post-training systems
\citep{xiao2026mimov2flashtechnicalreport,Nemotron_Cascade_2}.
Related methods adjust domain proportions during pretraining or multi-domain
reinforcement learning
\citep{xie2023doremi,liang2025modomodomultidomaindatamixtures}, but do not
coordinate teacher feedback on the same trajectory. Among recent MOPD methods,
Open-MOPD balances token shares and teacher--student gaps while refreshing stale
rewards \citep{gao2026openmopd}, and D$^3$-MOPD schedules domains by their
remaining distillation gaps and learning progress
\citep{sun2026d3mopdadaptivedynamicdomain}. Other methods separate
capability-recovery and domain-preservation updates
\citep{chen2026counteractionawaremultiteacheronpolicydistillation}, or calibrate
confidence across heterogeneous teachers
\citep{yin2026hopdconfidenceawareheterogeneous}. Gradient-conflict methods such
as PCGrad and conflict-averse gradient descent instead modify already-computed
task gradients \citep{yu2020gradient,NeurIPS2021-Liu}. None uses non-routed
teachers to evaluate the current target-domain trajectory. \method lets routed
and auxiliary teachers assess the same prefixes: their agreement guides
token-level credit assignment, while auxiliary supervision defines preventive
gradient constraints before domain updates are aggregated.

\section{Method}
\label{sec:method}


\begin{figure}[t]
    \centering
    \includegraphics[width=\textwidth]{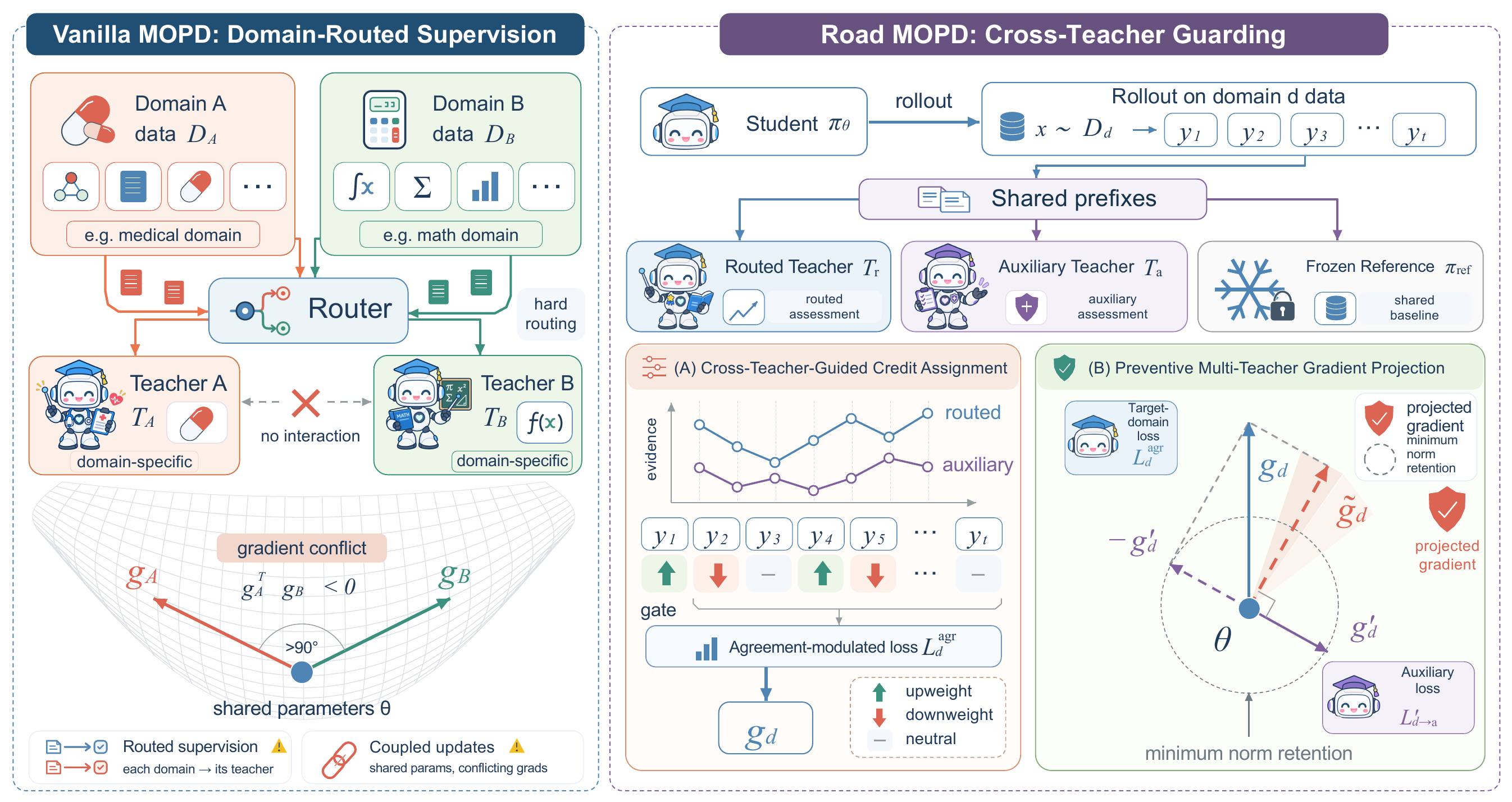}
    \caption{\textbf{Vanilla MOPD and \method.}
    Left: Vanilla MOPD's domain-routed supervision can produce conflicting
    updates to shared parameters.
    Right: \method uses same-trajectory auxiliary-teacher feedback for
    (A) agreement-guided token credit assignment and
    (B) preventive gradient projection under a
    target-gradient norm-retention constraint.}
    \label{fig:method-overview}
\end{figure}

\subsection{Problem Setup and On-Policy Distillation}
\label{sec:method:setup}

Let \(\pi_\theta\) be the student policy, \(\pi_T\) a teacher, and
\(\mathcal{D}\) a prompt distribution. For \(x\sim\mathcal{D}\), the
student samples \(y=(y_1,\ldots,y_T)\) with prefixes
\(h_t=(x,y_{<t})\). OPD minimizes the sequence-level reverse-KL objective:
\begin{equation}
    \mathcal{L}_{\mathrm{OPD}}(\theta;\pi_T,\mathcal{D})
    = \mathbb{E}_{x\sim\mathcal{D}}
      D_{\mathrm{KL}}\!\left(
        \pi_\theta(\cdot\mid x)\,\|\,\pi_T(\cdot\mid x)
      \right).
    \label{eq:opd}
\end{equation}

Following prior work~\citep{yang2026gopd}, we express OPD as
relative-reward maximization with a KL penalty to a frozen reference
\(\pi_{\mathrm{ref}}\). With reward scale \(\lambda>0\), we maximize
\begin{equation}
    \mathcal{J}_{\lambda}(\theta)
    =\mathbb{E}_\theta\!\left[
      \lambda\log\frac{\pi_T(y\mid x)}{\pi_{\mathrm{ref}}(y\mid x)}
      -D_{\mathrm{KL}}\!\left(
        \pi_\theta(\cdot\mid x)\,\|\,\pi_{\mathrm{ref}}(\cdot\mid x)
      \right)\right].
    \label{eq:lambda-opd}
\end{equation}
where \(\mathbb{E}_\theta\) averages over \(x\sim\mathcal{D}\) and
\(y\sim\pi_\theta(\cdot\mid x)\). Maximizing this objective recovers
standard OPD at \(\lambda=1\); \(0<\lambda<1\) attenuates the
teacher's relative reward, while \(\lambda>1\) amplifies it (ExOPD).
The reference provides a shared baseline, not an additional teacher.

For \(K\) domain--teacher pairs
\(\{(\mathcal{D}_d,\pi_{T_d})\}_{d=1}^{K}\), MOPD routes each prompt
to its domain teacher and minimizes the equal-weight objective:
\begin{equation}
    \mathcal{L}_{\mathrm{MOPD}}(\theta)
    =\frac{1}{K}\sum_{d=1}^{K}
      \mathcal{L}_{\mathrm{OPD}}(\theta;\pi_{T_d},\mathcal{D}_d).
    \label{eq:mopd}
\end{equation}
For a fixed target domain \(d\in\{1,\ldots,K\}\), denote its routed teacher by
\(T_{\mathrm r}=T_d\) and a teacher from an auxiliary domain by
\(T_{\mathrm a}\). The upright labels \(\mathrm r\) and \(\mathrm a\)
denote teacher roles, while \(d\) indexes domains. Both teachers
and the reference score the same student-generated tokens under teacher
forcing, without additional rollouts. We write \(\ell_t^{\mathrm r}\),
\(\ell_t^{\mathrm a}\), and \(\ell_t^{\mathrm{ref}}\) for their respective
log-probabilities of \(y_t\) given \(h_t\).

\subsection{Cross-Teacher-Guided Credit Assignment}
\label{sec:method:token}

A fixed \(\lambda\) in Eq.~(\ref{eq:lambda-opd}) applies the same
reward scale to every token, although a trajectory may contain
both transferable reasoning patterns and domain-specific
corrections. The routed teacher alone provides no cross-domain
check on these corrections. We therefore query an auxiliary
teacher on the same student-generated prefixes and use its
feedback to adjust the routed teacher's token-level reward.
This strengthens supervision supported by both teachers and
attenuates conflicting signals.
Importantly, the auxiliary teacher is not treated as a supervisor
for solving the target-domain task. Its relative preference is used
only as a compatibility signal for the capability represented by
that teacher.

To compare teacher preferences, we measure each teacher's
log-probability relative to the shared frozen reference.
Their token-wise agreement is
\begin{equation}
    s_t
    =\underbrace{\tanh\!\left(
       \frac{\ell_t^{\mathrm r}-\ell_t^{\mathrm{ref}}}{\tau}
     \right)}_{\text{routed-teacher evidence}}
     \times
     \underbrace{\tanh\!\left(
       \frac{\ell_t^{\mathrm a}-\ell_t^{\mathrm{ref}}}{\tau}
     \right)}_{\text{auxiliary-teacher evidence}},
    \qquad s_t\in[-1,1].
    \label{eq:teacher-reward-agreement}
\end{equation}
Positive \(s_t\) means that both teachers favor the same direction
of correction relative to the reference. Negative \(s_t\)
indicates opposing preferences, with values near \(-1\)
corresponding to strong cross-teacher disagreement. The score
approaches zero when either teacher provides little evidence
beyond the reference.

We use this agreement to replace the global reward scale in
Eq.~(\ref{eq:lambda-opd}) with a token-wise coefficient:
\begin{equation}
    \lambda_t
    =\operatorname{clip}\!\left(
       1+\rho_{\mathrm{g}}s_t,\lambda_{\min},\lambda_{\max}
     \right).
    \label{eq:token-wise-lambda}
\end{equation}
Here, \(\rho_{\mathrm g}>0\) controls the modulation strength;
the subscript \(\mathrm g\) denotes the agreement gate.
When \(s_t>0\), \(\lambda_t>1\) strengthens the routed teacher's
relative reward on tokens supported by both teachers. When
\(s_t<0\), \(\lambda_t<1\) attenuates the reward, with stronger
disagreement producing greater attenuation. When \(s_t=0\),
\(\lambda_t=1\) recovers standard OPD. Clipping limits both
amplification and attenuation.

For a sampled token \(y_t\) and a comparison policy \(q\), the
\(k_1\) log-ratio estimator is
\(k_{1,t}(\pi_\theta\|q)=\log\pi_\theta(y_t\mid h_t)-\log q(y_t\mid h_t)\).
The modulated signal
\(\lambda_t(\ell_t^{\mathrm r}-\ell_t^{\mathrm{ref}})-k_{1,t}(\pi_\theta\|\pi_{\mathrm{ref}})\)
is clipped and detached as the advantage in the PPO-style surrogate
\(\ell_{d,i,t}^{\mathrm{agr}}(\theta)\). Gradients flow through the policy
importance ratio, not the advantage (Appendix~\ref{app:distillation-objective}).
With multiple auxiliary teachers, we average their token-wise agreement
with the routed teacher before applying the same reward modulation
(Appendix~\ref{app:distillation-objective}).

\subsection{Preventive Multi-Teacher Gradient Projection}
\label{sec:method:gradient}

Token-level agreement does not guarantee compatible parameter
updates, since token supervision signals are aggregated through
shared parameters. We therefore formulate gradient correction as
a constrained optimization problem: minimize conflict with an
auxiliary teacher on the same trajectories while preserving a
minimum fraction of the target-gradient norm.

We first compute the target-domain gradient
\(g_d=\nabla_\theta\mathcal{L}_{d}^{\mathrm{agr}}\)
from the agreement-modulated loss. To construct the auxiliary
gradient, we replace the routed
teacher with the auxiliary teacher and disable agreement
modulation. Let
\(\ell_{d\rightarrow\mathrm a,i,t}^{\mathrm{OPD}}\)
denote the resulting plain token-level OPD loss. Applying the
same response-level reduction gives the auxiliary loss and gradient,
which we distinguish from their target-domain counterparts using primes:
\begin{equation}
    \mathcal{L}'_{d\rightarrow\mathrm a}
    =\frac{1}{|\mathcal B_d|}\sum_{i\in\mathcal B_d}\frac{1}{T_i}
      \sum_{t=1}^{T_i}\ell_{d\rightarrow\mathrm a,i,t}^{\mathrm{OPD}},
    \qquad
    g_d'
    =\nabla_\theta\mathcal{L}'_{d\rightarrow\mathrm a}.
    \label{eq:v2-auxiliary-gradient}
\end{equation}
Both losses use the same stored domain-\(d\) trajectories.
Thus, \(g_d'\) reflects the auxiliary teacher's preferences on
the current behavior, rather than its training gradient on
auxiliary-domain data. Differences in knowledge, response style,
or confidence can shape this signal; its usefulness as a
constraint is assessed through joint-domain performance.

We optimize the corrected gradient separately for each trainable
parameter tensor. Let \(g\) and \(g'\) be nonzero slices of
\(g_d\) and \(g_d'\), and let \(v\) denote a candidate corrected
gradient. For an intended tensor-level step \(-\eta v\) with \(\eta>0\), the
first-order change in the same-trajectory auxiliary loss is
\(-\eta\langle v,g'\rangle\). Thus, \(\langle g,g'\rangle<0\)
indicates local conflict; otherwise, we retain \(g\) unchanged.
For a conflicting pair, we restrict \(v\) to the line segment
between \(g\) and its full projection onto the hyperplane
orthogonal to \(g'\)~\citep{yu2020gradient}:
\begin{equation}
    g_{\mathrm{full}}
    =g-\frac{\langle g,g'\rangle}
            {\|g'\|_2^2}g',
    \qquad
    \mathcal S(g,g')=\operatorname{conv}\{g,g_{\mathrm{full}}\}.
    \label{eq:projection-feasible-set}
\end{equation}
Here, \(\operatorname{conv}\) denotes the convex hull.
This restriction preserves the component orthogonal to \(g'\)
and only removes the conflicting component, up to full projection.
Full projection eliminates local conflict but can nearly erase
\(g\) when the gradients oppose each other. We therefore choose
the gradient with the least residual conflict subject to a
norm-retention floor \(m\in(0,1]\):
\begin{equation}
    \widetilde g
    =\underset{v\in\mathcal S(g,g')}{\arg\min}\;
      -\langle v,g'\rangle
    \quad\text{subject to}\quad
    \|v\|_2^2\geq m^2\|g\|_2^2.
    \label{eq:projection-optimization}
\end{equation}
On this segment, \(-\langle v,g'\rangle\geq0\) measures the
predicted auxiliary-loss increase per unit step size, while the
constraint preserves the target-gradient norm. The solution can
be written as a partial projection:
\begin{equation}
    \widetilde g
    =g-\alpha^\star
      \frac{\langle g,g'\rangle}{\|g'\|_2^2}g',
    \qquad \alpha^\star\in[0,1].
    \label{eq:v2-projection}
\end{equation}
Here, \(\alpha^\star\) is the fraction of the conflicting component
removed. We use full projection (\(\alpha^\star=1\)) whenever it
satisfies the norm floor; otherwise, we remove as much conflict as
the floor permits, with \(\|\widetilde g\|_2=m\|g\|_2\).
For non-conflicting pairs, \(\alpha^\star=0\).
The closed-form coefficient and its derivation are given in
Appendix~\ref{app:projection-derivation}.
These are local statements about the same-trajectory auxiliary loss,
not guarantees of auxiliary-domain performance after an optimizer step.
Tensors with negligible gradient norm are left unchanged; auxiliary
gradients are never applied as separate optimizer updates.
The projected tensors form \(\widetilde g_d\).

With multiple auxiliary teachers, we apply the projections sequentially
under a shared norm-retention floor relative to the original target gradient
(Appendix~\ref{app:projection-derivation}). Method-specific
hyperparameters are summarized in Appendix~\ref{app:hyperparameters}.

A related approach in multi-task learning is
PCGrad~\citep{yu2020gradient}, which performs \textbf{\mbox{post-hoc}}
gradient surgery by subtracting conflicting components from
independently computed task gradients. Our projection is
\textbf{preventive}: the auxiliary teacher assesses the current
target-domain trajectory to construct a gradient constraint
specifically for that update, using its feedback to constrain potentially
incompatible corrections while retaining target-domain learning signals.
We empirically compare these two projection strategies in
Section~\ref{sec:experiments:analysis}.

\subsection{Overall Objective and Training}
\label{sec:method:overall}

We use sequence-mean aggregation, averaging token losses
within each response and then across responses and domains. Auxiliary
losses use the same reduction but guide only the gradient correction,
so the overall distillation objective includes only target-domain losses:
\begin{equation}
    \mathcal{L}_{\mathrm{overall}}(\theta)
    =\frac{1}{K}\sum_{d=1}^{K}
      \frac{1}{|\mathcal B_d|}
      \sum_{i\in\mathcal B_d}\frac{1}{T_i}
      \sum_{t=1}^{T_i}\ell_{d,i,t}^{\mathrm{agr}}(\theta).
    \label{eq:overall-seqmean}
\end{equation}

After applying the preventive projection in
Section~\ref{sec:method:gradient}, we average only the projected target-domain
gradients, using the same domain weights as Eq.~(\ref{eq:overall-seqmean}).
The resulting \(g_{\mathrm{update}}\) is globally clipped for one AdamW step:
\begin{equation}
    g_{\mathrm{update}}
    =\frac{1}{K}\sum_{d=1}^{K}\widetilde g_d.
    \label{eq:merged-update}
\end{equation}

\section{Experiments}
\label{sec:experiments}

\providecommand{\resultblank}{\phantom{00.00}}
\definecolor{resultsetting}{RGB}{232,239,243}
\definecolor{resultavg}{RGB}{244,244,244}
\definecolor{resultconflict}{RGB}{156,68,68}
\newcommand{\bestscore}[1]{\textbf{#1}}
\newcommand{\secondscore}[1]{\underline{#1}}
\newcommand{\settingheader}[1]{%
    \begin{tabular}{@{}p{\textwidth}@{}}
        \specialrule{\heavyrulewidth}{0pt}{0pt}
        \rowcolor{resultsetting}[0pt][0pt]
        \rule{0pt}{2.2ex}{\small\textbf{#1}} \\
        \specialrule{\lightrulewidth}{0pt}{0pt}
    \end{tabular}\par\nointerlineskip
}

\newcommand{\modelheader}[1]{%
    \begin{tabular}{@{}p{\textwidth}@{}}
        \rowcolor{resultavg}[0pt][0pt]
        \rule{0pt}{2.1ex}{\fontsize{8}{9.5}\selectfont\textbf{#1}} \\
    \end{tabular}\par\nointerlineskip
}
\newcommand{\qwenSmallMathMedical}{%
    \resizebox{\textwidth}{!}{%
    \begin{tabular}{@{}l*{4}{c}>{\columncolor{resultavg}[2pt][2pt]}c*{4}{c}>{\columncolor{resultavg}[2pt][2pt]}c>{\columncolor{resultavg}[2pt][2pt]}c@{}}
        \multirow{2}{*}{\textbf{Method}}
        & \multicolumn{5}{c}{\textbf{Math}}
        & \multicolumn{5}{c}{\textbf{Medical}}
        & \multicolumn{1}{c}{\cellcolor{white}\textbf{All}} \\
        \cmidrule(lr){2-6}\cmidrule(lr){7-11}\cmidrule(l){12-12}
        & AIME24 & AIME25 & \shortstack{HMMT25\\Feb.} & \shortstack{HMMT25\\Nov.} & \shortstack{Domain\\Avg.} & MedQA & \shortstack{MedXpert\\QA} & MedMCQA & \shortstack{SuperGPQA\\-Med} & \shortstack{Domain\\Avg.} & \raisebox{0.5\normalbaselineskip}{\textit{Avg.}} \\
        \midrule
        Student & 11.98 & 10.21 & 6.04 & 3.96 & 8.05 & 46.03 & 10.33 & 42.55 & 22.58 & 30.37 & 19.21 \\
        Single OPD & 39.90 & 34.48 & 18.75 & 17.29 & 27.60 & 53.81 & 13.39 & 50.99 & 24.50 & 35.67 & 31.64 \\
        \midrule
        Vanilla MOPD & \secondscore{41.46} & \secondscore{35.10} & 18.85 & 18.02 & \secondscore{28.36} & 47.37 & \bestscore{11.39} & 45.25 & 20.69 & 31.18 & 29.77 \\
        ExOPD & \bestscore{41.56} & \bestscore{35.42} & 18.13 & \secondscore{18.65} & \bestscore{28.44} & \secondscore{49.88} & 9.80 & \secondscore{49.49} & \secondscore{22.87} & \secondscore{33.01} & \secondscore{30.72} \\
        Open-MOPD & 37.40 & 31.98 & \secondscore{18.96} & 18.33 & 26.67 & 46.58 & \bestscore{11.39} & 44.20 & 20.04 & 30.55 & 28.61 \\
        CaMOPD & 36.35 & 33.13 & \bestscore{19.06} & 18.23 & 26.69 & 47.84 & 10.16 & 44.63 & 21.49 & 31.03 & 28.86 \\
        \textbf{Ours} & 40.10 & 31.88 & 17.81 & \bestscore{19.06} & 27.21 & \bestscore{58.13} & \secondscore{11.06} & \bestscore{52.35} & \bestscore{26.03} & \bestscore{36.89} & \bestscore{32.05} \\
        \bottomrule
    \end{tabular}%
    }\par\nointerlineskip
}

\newcommand{\qwenSmallCodeMedical}{%
    \resizebox{\textwidth}{!}{%
    \begin{tabular}{@{}l*{3}{c}>{\columncolor{resultavg}[2pt][2pt]}c*{4}{c}>{\columncolor{resultavg}[2pt][2pt]}c>{\columncolor{resultavg}[2pt][2pt]}c@{}}
        \multirow{2}{*}{\textbf{Method}}
        & \multicolumn{4}{c}{\textbf{Code}}
        & \multicolumn{5}{c}{\textbf{Medical}}
        & \multicolumn{1}{c}{\cellcolor{white}\textbf{All}} \\
        \cmidrule(lr){2-5}\cmidrule(lr){6-10}\cmidrule(l){11-11}
         & HumanEval+ & MBPP+ & LCB v6 & \shortstack{Domain\\Avg.} & MedQA & \shortstack{MedXpert\\QA} & MedMCQA & \shortstack{SuperGPQA\\-Med} & \shortstack{Domain\\Avg.} & \raisebox{0.5\normalbaselineskip}{\textit{Avg.}} \\
        \midrule
        Student & 60.37 & 53.44 & 12.57 & 42.13 & 46.03 & 10.33 & 42.55 & 22.58 & 30.37 & 36.25 \\
        Single OPD & 73.48 & 59.13 & 22.43 & 51.68 & 53.81 & 13.39 & 50.99 & 24.50 & 35.67 & 43.67 \\
        \midrule
        Vanilla MOPD & 71.95 & 58.27 & \secondscore{23.00} & 51.07 & 46.11 & \secondscore{11.02} & 45.68 & 20.91 & 30.93 & 41.00 \\
        ExOPD & \bestscore{75.76} & \bestscore{62.30} & \bestscore{24.43} & \bestscore{54.16} & 44.62 & 10.37 & 44.85 & 19.24 & 29.77 & \secondscore{41.97} \\
        Open-MOPD & 71.95 & 58.60 & 20.43 & 50.33 & 45.80 & 10.78 & 43.46 & 19.64 & 29.92 & 40.12 \\
        CaMOPD & 71.65 & 55.29 & \secondscore{23.00} & 49.98 & \secondscore{47.53} & 10.33 & \secondscore{45.80} & \secondscore{21.71} & \secondscore{31.34} & 40.66 \\
        \textbf{Ours} & \secondscore{73.78} & \secondscore{61.44} & \bestscore{24.43} & \secondscore{53.22} & \bestscore{56.56} & \bestscore{12.24} & \bestscore{51.35} & \bestscore{25.88} & \bestscore{36.51} & \bestscore{44.86} \\
        \bottomrule
    \end{tabular}%
    }\par\nointerlineskip
}

\begin{table}[t]
    \caption{\textbf{Main results across paired-domain settings.}
    Domain Avg. averages benchmarks within each domain; Avg. averages the two domains.
    Bold/underlined scores mark the best/second-best multi-domain methods per student model.}
    \label{tab:main-results}
    \centering
    \renewcommand{\arraystretch}{1.08}

    \settingheader{(a) Math $\times$ Medical}
    \modelheader{Student Model: Qwen3-4B}
    \resizebox{\textwidth}{!}{%
    \begin{tabular}{@{}l*{4}{c}>{\columncolor{resultavg}[2pt][2pt]}c*{4}{c}>{\columncolor{resultavg}[2pt][2pt]}c>{\columncolor{resultavg}[2pt][2pt]}c@{}}
        \multirow{2}{*}{\textbf{Method}}
        & \multicolumn{5}{c}{\textbf{Math}}
        & \multicolumn{5}{c}{\textbf{Medical}}
        & \multicolumn{1}{c}{\cellcolor{white}\textbf{All}} \\
        \cmidrule(lr){2-6}\cmidrule(lr){7-11}\cmidrule(l){12-12}
        & AIME24 & AIME25 & \shortstack{HMMT25\\Feb.} & \shortstack{HMMT25\\Nov.} & \shortstack{Domain\\Avg.} & MedQA & \shortstack{MedXpert\\QA} & MedMCQA & \shortstack{SuperGPQA\\-Med} & \shortstack{Domain\\Avg.} & \raisebox{0.5\normalbaselineskip}{\textit{Avg.}} \\
        \midrule
        Student      & 21.46 & 21.77 & 10.00 & 8.23 & 15.37 & 64.73 & 13.80 & 55.63 & 29.87 & 41.01 & 28.19 \\
        Single OPD   & 58.13 & 51.67 & 31.56 & 35.73 & 44.27 & 77.06 & 15.31 & 62.44 & 36.81 & 47.90 & 46.09 \\
        \midrule
        Vanilla MOPD & 58.75 & \secondscore{55.21} & 31.98 & 37.92 & 45.97 & 62.61 & 12.24 & 55.39 & 27.55 & 39.45 & 42.71 \\
        ExOPD        & \secondscore{61.67} & \bestscore{55.62} & \secondscore{33.54} & 37.92 & \secondscore{47.19} & 69.99 & 13.76 & 59.50 & 32.23 & 43.87 & 45.53 \\
        Open-MOPD    & 56.56 & 55.10 & 32.60 & 37.08 & 45.34 & 70.31 & 13.06 & 60.94 & 33.76 & 44.52 & 44.93 \\
        CaMOPD       & 61.15 & 54.27 & 33.12 & \secondscore{38.02} & 46.64 & \bestscore{77.45} & \bestscore{16.61} & \secondscore{61.89} & \secondscore{35.03} & \secondscore{47.75} & \secondscore{47.20} \\
        \textbf{Ours} & \bestscore{63.85} & 53.96 & \bestscore{33.65} & \bestscore{38.96} & \bestscore{47.60} & \secondscore{77.06} & \secondscore{15.67} & \bestscore{62.83} & \bestscore{37.35} & \bestscore{48.23} & \bestscore{47.92} \\
        \bottomrule
    \end{tabular}%
    }\par\nointerlineskip
    \modelheader{Student Model: Qwen3-1.7B}
    \qwenSmallMathMedical
    \settingheader{(b) Code $\times$ Medical}
    \modelheader{Student Model: Qwen3-4B}
    \resizebox{\textwidth}{!}{%
    \begin{tabular}{@{}l*{3}{c}>{\columncolor{resultavg}[2pt][2pt]}c*{4}{c}>{\columncolor{resultavg}[2pt][2pt]}c>{\columncolor{resultavg}[2pt][2pt]}c@{}}
        \multirow{2}{*}{\textbf{Method}}
        & \multicolumn{4}{c}{\textbf{Code}}
        & \multicolumn{5}{c}{\textbf{Medical}}
        & \multicolumn{1}{c}{\cellcolor{white}\textbf{All}} \\
        \cmidrule(lr){2-5}\cmidrule(lr){6-10}\cmidrule(l){11-11}
         & HumanEval+ & MBPP+ & LCB & \shortstack{Domain\\Avg.} & MedQA & \shortstack{MedXpert\\QA} & MedMCQA & \shortstack{SuperGPQA\\-Med} & \shortstack{Domain\\Avg.} & \raisebox{0.5\normalbaselineskip}{\textit{Avg.}} \\
        \midrule
        Student & 76.68 & 65.08 & 17.14 & 52.97 & 64.73 & 13.80 & 55.63 & 29.87 & 41.01 & 46.99 \\
        Single OPD & 84.30 & 70.44 & 28.00 & 60.91 & 77.06 & 15.31 & 62.44 & 36.81 & 47.90 & 54.41 \\
        \midrule
        Vanilla MOPD & 84.76 & 69.11 & 27.71 & 60.53 & 64.96 & 12.78 & 55.63 & 29.62 & 40.75 & 50.64 \\
        ExOPD & \secondscore{85.98} & 68.98 & \bestscore{30.00} & \secondscore{61.65} & 72.82 & 13.63 & 60.46 & 32.81 & 44.93 & 53.29 \\
        Open-MOPD & 84.45 & \secondscore{70.77} & 27.29 & 60.83 & \bestscore{76.83} & \secondscore{15.59} & 61.68 & \secondscore{35.25} & \secondscore{47.34} & \secondscore{54.09} \\
        CaMOPD & 84.91 & 69.58 & 27.43 & 60.64 & \secondscore{76.12} & 15.39 & \secondscore{62.49} & 35.10 & 47.28 & 53.96 \\
        \textbf{Ours} & \bestscore{87.65} & \bestscore{72.69} & \secondscore{28.71} & \bestscore{63.02} & 75.73 & \bestscore{15.96} & \bestscore{62.97} & \bestscore{36.33} & \bestscore{47.75} & \bestscore{55.38} \\
        \bottomrule
    \end{tabular}%
    }\par\nointerlineskip
    \modelheader{Student Model: Qwen3-1.7B}
    \qwenSmallCodeMedical
\end{table}

\subsection{Experimental Setup}
\label{sec:experiments:setup}

\noindent\textbf{Evaluation datasets.}
We evaluate 11 benchmarks across Code (HumanEval+, MBPP+, and LiveCodeBench)
\citep{liu2023evalplus,jain2024livecodebench}, Math (AIME 2024/2025 and HMMT February/November 2025)
\citep{maa2024aime,maa2025aime,hmmt2025feb,hmmt2025nov}, and Medical
(MedQA, MedXpertQA, MedMCQA, and SuperGPQA-Med)
\citep{jin2020medqa,zuo2025medxpertqa,pal2022medmcqa,mapteam2025supergpqa}.
We report Avg@4 accuracy for Code, Avg@32 for Math, and Avg@1 for Medical.
Avg@$n$ averages correctness over $n$ completions per problem and then over
problems, equivalent to the same-sample pass@1 estimate for Math and Code.

\noindent\textbf{Baselines.}
We compare with four representative approaches:
(1) Vanilla MOPD routes each on-policy rollout only to its corresponding domain teacher and applies
token-level reverse-KL supervision \citep{ma2026mopdmultiteacheronpolicydistillation};
(2) ExOPD constructs an extrapolated teacher target using the frozen student initialization as a
reference \citep{yang2026gopd};
(3) CaMOPD separates capability-recovery and domain-preservation updates and prioritizes samples with
larger teacher--student gaps \citep{chen2026counteractionawaremultiteacheronpolicydistillation}; and
(4) Open-MOPD mitigates cross-domain imbalance through token-share balancing, dynamic domain loss weighting,
and reward refresh \citep{gao2026openmopd}.

\noindent\textbf{Implementation details.}
We use Qwen3-4B and Qwen3-1.7B students with thinking disabled
\citep{qwen2025qwen3}. Math and Code teachers use specialized Qwen3-4B checkpoints
\citep{yang2026gopd}.
The Medical teacher is II-Medical-8B, a Qwen3-8B RL checkpoint
\citep{intelligentinternet2025iimedical8b}. Each paired setting uses 10,000 sampled
prompts per domain from DeepMath, Eurus-RL-Code, and Medical sources including
the MedMCQA training split (Table~\ref{tab:app-training-data}).
Full configurations are in Appendix~\ref{app:experimental-setup}.

\begin{samepage}
\subsection{Main Results}
\label{sec:experiments:main}

\begin{table}[t]
    \begin{minipage}[t]{0.441\textwidth}
        \vspace{0pt}
        \caption{\raggedright\textbf{Low-conflict generalization: \mbox{Math $\times$ Code.}} Qwen3-4B domain averages (\%) and their mean (All).}
        \label{tab:low-conflict-results}
    \end{minipage}\hfill
    \begin{minipage}[t]{0.539\textwidth}
        \vspace{0pt}
        \caption{\raggedright\textbf{Three-domain generalization: \mbox{Math $\times$ Code $\times$ Medical.}} Qwen3-4B domain averages (\%) and their mean (All).}
        \label{tab:three-domain-results}
    \end{minipage}
    \par\nointerlineskip
    \begin{minipage}[t]{0.441\textwidth}
        \vspace{0pt}
\centering
\fontsize{8}{9.6}\selectfont
\setlength{\tabcolsep}{6pt}
\renewcommand{\arraystretch}{1.12}
\begin{tabular*}{\linewidth}{@{\extracolsep{\fill}}lcc>{\columncolor{resultavg}[2pt][2pt]}c@{}}
    \toprule
    \textbf{Method} & \textbf{Math} & \textbf{Code} & \multicolumn{1}{c@{}}{\cellcolor{white}\textbf{All}} \\
    \midrule
    Student & 15.37 & 52.97 & 34.17 \\
    Single OPD & 44.27 & 60.91 & 52.59 \\
    \midrule
    Vanilla MOPD & 46.12 & 61.24 & 53.68 \\
    ExOPD & \bestscore{48.28} & \secondscore{61.82} & \secondscore{55.05} \\
    Open-MOPD & 44.32 & 59.82 & 52.07 \\
    CaMOPD & 46.30 & 60.37 & 53.34 \\
    \textbf{Ours} & \secondscore{47.73} & \bestscore{63.07} & \bestscore{55.40} \\
    \bottomrule
\end{tabular*}\par

    \end{minipage}\hfill
    \begin{minipage}[t]{0.539\textwidth}
        \vspace{0pt}
\centering
\fontsize{8}{9.6}\selectfont
\setlength{\tabcolsep}{6pt}
\renewcommand{\arraystretch}{1.12}
\begin{tabular*}{\linewidth}{@{\extracolsep{\fill}}lccc>{\columncolor{resultavg}[2pt][2pt]}c@{}}
    \toprule
    \textbf{Method} & \textbf{Math} & \textbf{Code} & \textbf{Medical} & \multicolumn{1}{c@{}}{\cellcolor{white}\textbf{All}} \\
    \midrule
    Student & 15.37 & 52.97 & 41.01 & 36.45 \\
    Single OPD & 44.27 & 60.91 & 47.90 & 51.03 \\
    \midrule
    Vanilla MOPD & 46.98 & 61.26 & 41.03 & 49.76 \\
    ExOPD & \secondscore{47.09} & \secondscore{61.52} & 45.62 & 51.41 \\
    Open-MOPD & 44.06 & 60.56 & 38.83 & 47.82 \\
    CaMOPD & 45.96 & 60.97 & \secondscore{48.20} & \secondscore{51.71} \\
    \textbf{Ours} & \bestscore{47.29} & \bestscore{62.94} & \bestscore{48.45} & \bestscore{52.89} \\
    \bottomrule
\end{tabular*}\par

    \end{minipage}
\end{table}

\noindent\textbf{Best overall performance across domain pairs and student sizes.}
Across Math $\times$ Medical and Code $\times$ Medical with both student sizes,
\method achieves the highest two-domain average among the compared
multi-domain methods in all four settings (Table~\ref{tab:main-results}),
exceeding the strongest overall baseline in each by 0.72--2.89 points.

\end{samepage}

\noindent\textbf{4B students: Medical recovery accompanies gains in the other domain.}
In both Math $\times$ Medical and Code $\times$ Medical, Vanilla MOPD lowers
Medical accuracy below the initial student's 41.01, while ExOPD, Open-MOPD,
and CaMOPD mitigate this degradation to varying degrees.
\method reaches Medical averages of 48.23 and 47.75, respectively, recovering
to at or near the Single OPD reference of 47.90 and exceeding the strongest
multi-domain Medical baseline in each setting by 0.48 and 0.41 points.
It also achieves the highest Math and Code averages in the respective
settings, indicating that Medical recovery accompanies gains in the other
domain's average performance.

\begin{figure}[t]
    \centering
    \includegraphics[width=\textwidth]{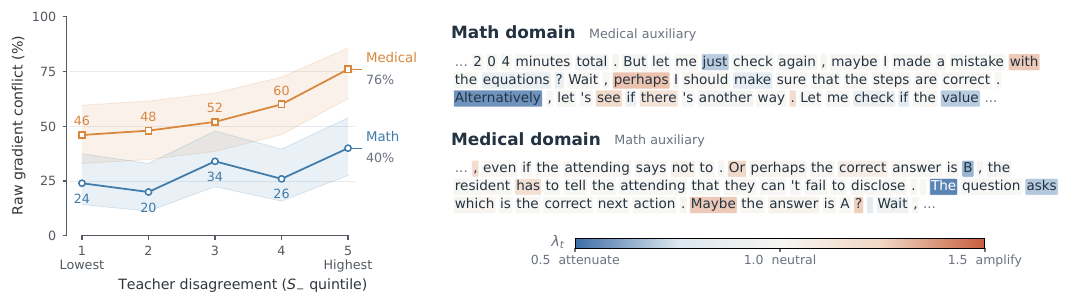}
    \caption{\textbf{Teacher disagreement and local supervision.}
    Left: raw gradient conflict by disagreement quintile
    (Math: blue; Medical: orange).
    Right: Math (top) and Medical (bottom) response windows; blue attenuates
    ($\lambda_t<1$) and orange amplifies ($\lambda_t>1$) token supervision.}
    \label{fig:cross-teacher-advantage-modulation}
\end{figure}

\noindent\textbf{1.7B students: Larger Medical gains improve the cross-domain trade-off.}
With the smaller student, all multi-domain baselines fall below the Single OPD
Medical average of 35.67 in both pairs. \method instead reaches 36.89 and
36.51, respectively, exceeding this reference in both settings. Its margins
over the strongest multi-domain Medical baseline in each setting increase
to 3.88 and 5.17 points. Although its Math and Code averages trail ExOPD by
1.23 and 0.94 points, respectively, the Medical gains yield two-domain
averages that exceed ExOPD by 1.33 and 2.89 points, demonstrating a better
overall result when domains are weighted equally.

\subsection{Further Analysis}
\label{sec:experiments:analysis}

\noindent\textbf{Generalizing to lower-conflict settings and more domains.}
Tables~\ref{tab:low-conflict-results} and~\ref{tab:three-domain-results}
show that \method also improves joint performance in low-conflict and
three-domain settings. For Math $\times$ Code, it achieves the best Code
and second-best Math averages among the compared multi-domain methods.
Although its Math average trails ExOPD by 0.55 points, its 1.25-point Code
gain yields the highest All Avg. of 55.40, exceeding ExOPD by 0.35 points.
With all three domains, \method leads in every domain and achieves the
best All Avg. of 52.89, surpassing CaMOPD by 1.18 points.
This setting also tests multiple auxiliary teachers, using aggregated agreement
for credit assignment and sequential constraints for gradient protection.
Full per-benchmark results are in
Appendix~\ref{app:low-conflict-results}
(Tables~\ref{tab:app-low-conflict-results} and~\ref{tab:app-three-domain-results}).

\newsavebox{\pcgradplotbox}
\newsavebox{\pcgradcaptionbox}
\newsavebox{\ablationcaptionbox}
\newsavebox{\ablationbodybox}
\newlength{\pcgradablationheight}
\newlength{\ablationrowgap}
\begin{table}[t]
\centering
\sbox{\pcgradplotbox}{%
    \includegraphics[width=0.53\textwidth]{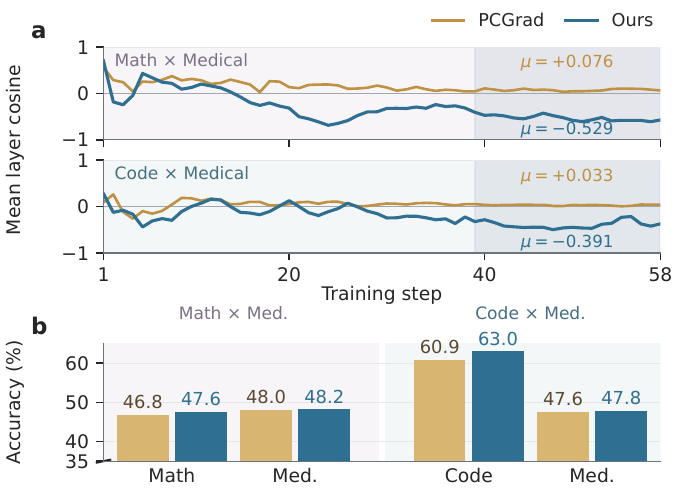}%
}
\sbox{\pcgradcaptionbox}{%
\begin{minipage}{0.53\textwidth}
    \setlength{\abovecaptionskip}{0pt}
    \captionof{figure}{\raggedright\textbf{Preventive projection vs.\ PCGrad.}
    (a) Mean layerwise cosine; labels average shaded steps 39--58.
    (b) Domain accuracy (\%).}
    \label{fig:cross-pcgrad-accuracy-layer-cosine}
\end{minipage}%
}
\sbox{\ablationcaptionbox}{%
\begin{minipage}{0.45\textwidth}
    \setlength{\abovecaptionskip}{0pt}
    \caption{\raggedright\textbf{Ablations of \method.}\newline
    Domain-average accuracy (\%).\newline
    Avg.: mean across the two domains.}
    \label{tab:ablation-results-v2}
\end{minipage}%
}
\newcommand{\ablationtablebody}{%
\begin{minipage}{0.45\textwidth}
    \centering
    \footnotesize
    \setlength{\tabcolsep}{3pt}
    \renewcommand{\arraystretch}{1.10}
    \begin{tabular*}{\linewidth}{@{\extracolsep{\fill}}lcc>{\columncolor{resultavg}[3pt][3pt]}c@{}}
        \toprule
        \textbf{Math $\times$ Med.}
        & \textbf{Math} & \textbf{Med.} & \textbf{Avg.} \\[\ablationrowgap]
        \midrule
        w/o Credit Assign.
        & 45.78 & 47.74 & 46.76 \\[\ablationrowgap]
        w/o Grad. Protect.
        & \secondscore{47.19} & 48.11 & \secondscore{47.65} \\[\ablationrowgap]
        w/o Adaptive Ret.
        & 45.78 & \bestscore{48.40} & 47.09 \\[\ablationrowgap]
        \midrule
        \method
        & \bestscore{47.60} & \secondscore{48.23} & \bestscore{47.92} \\[\ablationrowgap]
        \midrule
        \textbf{Code $\times$ Med.}
        & \textbf{Code} & \textbf{Med.} & \textbf{Avg.} \\[\ablationrowgap]
        \midrule
        w/o Credit Assign.
        & 59.54 & \secondscore{47.46} & 53.50 \\[\ablationrowgap]
        w/o Grad. Protect.
        & \secondscore{63.01} & 46.87 & \secondscore{54.94} \\[\ablationrowgap]
        w/o Adaptive Ret.
        & 62.85 & 47.03 & \secondscore{54.94} \\[\ablationrowgap]
        \midrule
        \method
        & \bestscore{63.02} & \bestscore{47.75} & \bestscore{55.38} \\[\ablationrowgap]
        \bottomrule
    \end{tabular*}
\end{minipage}%
}
\setlength{\ablationrowgap}{0pt}
\sbox{\ablationbodybox}{\ablationtablebody}
\setlength{\pcgradablationheight}{%
    \dimexpr\ht\pcgradplotbox+\dp\pcgradplotbox+
    \ht\pcgradcaptionbox+\dp\pcgradcaptionbox+3pt\relax}
\ifdim\dimexpr\ht\ablationcaptionbox+\dp\ablationcaptionbox+
    \ht\ablationbodybox+\dp\ablationbodybox+3pt\relax>\pcgradablationheight
    \setlength{\pcgradablationheight}{%
        \dimexpr\ht\ablationcaptionbox+\dp\ablationcaptionbox+
        \ht\ablationbodybox+\dp\ablationbodybox+3pt\relax}
\fi
\setlength{\ablationrowgap}{\dimexpr\pcgradablationheight-
    \ht\ablationcaptionbox-\dp\ablationcaptionbox-3pt-
    \ht\ablationbodybox-\dp\ablationbodybox\relax}
\divide\ablationrowgap by 10
\ifdim\ablationrowgap>0pt
    \sbox{\ablationbodybox}{\ablationtablebody}
\fi
\begin{minipage}[t][\pcgradablationheight][s]{0.53\textwidth}
    \vspace{0pt}
    \noindent\usebox{\pcgradplotbox}\par\nointerlineskip
    \vspace{3pt}
    \noindent\usebox{\pcgradcaptionbox}\par
    \vfill
    \vspace{0pt}
\end{minipage}\hfill
\begin{minipage}[t][\pcgradablationheight][s]{0.45\textwidth}
    \vspace{0pt}
    \noindent\usebox{\ablationcaptionbox}\par\nointerlineskip
    \vspace{3pt}
    \noindent\usebox{\ablationbodybox}\par
    \vfill
    \vspace{0pt}
\end{minipage}
\par
\end{table}

\noindent\textbf{Cross-teacher disagreement is associated with gradient conflict, supporting token-level credit assignment.}
To examine whether cross-teacher agreement reflects update compatibility,
we sample response batches at a fixed student checkpoint, averaging
$\max(-s_t,0)$ (Section~\ref{sec:method:token}) over all valid tokens per batch
to obtain disagreement $S_-$. Within each domain, we sort batches by $S_-$
from low to high into five equal-sized groups.
Figure~\ref{fig:cross-teacher-advantage-modulation}
(left) plots each group's fraction of batches with negative cosine between
raw full-parameter OPD gradients under the two teachers on identical responses.
Conflict rates trend upward from 24\% to 40\% in Math and 46\% to 76\% in
Medical. This suggests that teacher disagreement can signal potential update
conflicts, providing empirical support for agreement-guided credit assignment.
Figure~\ref{fig:cross-teacher-advantage-modulation} (right) visualizes token-level
credit assignment: relative to the shared reference, opposing teacher corrections on
\mbox{\emph{Alternatively}} in Math and \emph{The} in Medical attenuate the
routed teacher's relative reward signal, whereas aligned corrections on
\emph{perhaps} and \emph{Maybe} in the respective domains amplify it.
Probe details are provided in Appendix~\ref{app:evaluation-protocols};
full prompts and visualizations of extended responses
appear in Appendix~\ref{app:extended-token-visualizations}
(Figure~\ref{fig:app-extended-token-examples}).

\noindent\textbf{Preventive projection improves joint learning over post-hoc correction.}
Figure~\ref{fig:cross-pcgrad-accuracy-layer-cosine} compares preventive projection
with PCGrad under matched training settings; both retain cross-teacher credit
assignment and the same norm-retention mechanism.
PCGrad corrects conflicts between gradients computed independently on each
domain's own trajectories. Our method instead constrains target-domain updates
using auxiliary-teacher gradients on the same rollout.
This directly tests compatibility between the target gradient and
auxiliary-teacher feedback on the same behavior.
Unlike PCGrad's comparison of independent domain gradients, this constraint
can address local conflicts that may persist even when those gradients align.
Figure~\ref{fig:cross-pcgrad-accuracy-layer-cosine}a shows negative mean layerwise
cosines for same-trajectory pairs during much of training, whereas independently
formed domain-gradient pairs are mostly non-negative.
These patterns reflect different local compatibility criteria and do not
directly measure capability damage.
Across Math $\times$ Medical and Code $\times$ Medical, \method improves all four
domain-average accuracies over PCGrad by 0.18--2.12 percentage points
(Figure~\ref{fig:cross-pcgrad-accuracy-layer-cosine}b).
These gains support the utility of same-trajectory constraints for joint learning.

\FloatBarrier
\enlargethispage*{0pt}
\noindent\textbf{Each component improves joint performance.}
\label{sec:experiments:ablation}
Table~\ref{tab:ablation-results-v2} shows that removing either credit assignment
or gradient protection lowers all four domain averages across the two domain pairs.
Without credit assignment,
Math and Code accuracy drop by 1.82 and 3.48 points, respectively; removing gradient protection
reduces Medical accuracy by 0.88 points in Code $\times$ Medical.
Replacing adaptive retention with full projection lowers the two-domain mean
by 0.83 and 0.44 points
in Math $\times$ Medical and Code $\times$ Medical, respectively, despite a
0.17-point Medical gain in the former. The full method achieves the highest
equally weighted two-domain mean in both settings.
These results support adaptive retention as a way to balance conflict reduction
with target-domain learning, alongside credit assignment and gradient protection.
\FloatBarrier

\section{Conclusion}

We identify update incompatibility as a central challenge in multi-teacher
on-policy distillation: domain-routed supervision can produce conflicting updates
to shared student parameters. \method addresses this challenge using auxiliary
teachers to assess the same student trajectories. Agreement-aware token weighting
and preventive gradient projection suppress conflicting supervision and updates
while retaining target-domain learning signals. Across mathematics, medicine, and code,
\method mitigates high-conflict degradation, retains gains in compatible settings,
and achieves the best distilled-model average in each domain under joint
three-domain training. Our evaluation is limited to three text-based reasoning
domains, and auxiliary scoring and gradient computation add training cost.
Extending to broader domains and larger teacher pools calls for efficient teacher
selection and compatibility estimation.


\subsection*{AI use statement}

We used generative AI tools to assist with English translation, language
polishing, and the presentation of figures and tables. This assistance included
refining wording, improving clarity, and adjusting captions and layouts.
The authors retain full responsibility for the paper's technical content,
results, citations, and conclusions, as well as the accuracy and
integrity of the final manuscript.


\bibliography{iclr2027_conference}

@inproceedings{agarwal2024OPD,
  title = {{On-Policy Distillation of Language Models: Learning from Self-Generated Mistakes}},
  author = {Rishabh Agarwal
    and Nino Vieillard
    and Yongchao Zhou
    and Piotr Stanczyk
    and Sabela Ramos Garea
    and Matthieu Geist
    and Olivier Bachem},
  year = {2024},
  booktitle = {The Twelfth International Conference on Learning Representations, {ICLR} 2024, Vienna, Austria, May 7-11, 2024},
  url = {https://proceedings.iclr.cc/paper_files/paper/2024/hash/5be69a584901a26c521c2b51e40a4c20-Abstract-Conference.html},
  eprint = {2306.13649},
  archiveprefix = {arXiv},
}

@article{chen2024huatuogpto1,
  title         = {{HuatuoGPT-o1}, Towards Medical Complex Reasoning with {LLMs}},
  author        = {Junying Chen and Zhenyang Cai and Ke Ji and Xidong Wang and Wanlong Liu and Rongsheng Wang and Jianye Hou and Benyou Wang},
  journal={arXiv preprint arXiv:2412.18925},
  year          = {2024}
}

@article{chen2026counteractionawaremultiteacheronpolicydistillation,
  title={Counteraction-Aware Multi-Teacher On-Policy Distillation for General Capability Recovery with Domain Preservation},
  author={Tianlei Chen and Jiao Ou and Ziyuan Liu and Ruiming Tang and Jian Liang and Han Li},
  journal={arXiv preprint arXiv:2605.27115},
  year={2026}
}

@article{cui2025process,
  title = {{Process Reinforcement through Implicit Rewards}},
  author = {Cui, Ganqu
    and Yuan, Lifan
    and Wang, Zefan
    and Wang, Hanbin
    and Zhang, Yuchen
    and Chen, Jiacheng
    and others},
  journal = {arXiv preprint arXiv:2502.01456},
  year = {2025},
  url = {https://arxiv.org/abs/2502.01456},
  eprint = {2502.01456},
  archiveprefix = {arXiv},
}

@article{guo2025deepseekr1,
  title = {{DeepSeek-R1: Incentivizing Reasoning Capability in LLMs via Reinforcement Learning}},
  author = {{DeepSeek-AI}},
  year = {2025},
  eprint = {2501.12948},
  archiveprefix = {arXiv},
  primaryclass = {cs.CL},
  url = {https://arxiv.org/abs/2501.12948},
  journal = {arXiv preprint arXiv:2501.12948},
}

@article{xu2026deepseekv4,
  title = {{DeepSeek-V4: Towards Highly Efficient Million-Token Context Intelligence}},
  author = {{DeepSeek-AI}
    and Anyi Xu
    and Bangcai Lin
    and Bing Xue
    and Bingxuan Wang
    and Bingzheng Xu
    and others},
  year = {2026},
  eprint = {2606.19348},
  archiveprefix = {arXiv},
  primaryclass = {cs.CL},
  url = {https://arxiv.org/abs/2606.19348},
  journal = {arXiv preprint arXiv:2606.19348},
}

@misc{freedomintelligence2025medicalr1,
  author = {{FreedomIntelligence}},
  title = {{Medical-R1-Distill-Data}},
  year = {2025},
  howpublished = {Hugging Face dataset},
  url = {https://huggingface.co/datasets/FreedomIntelligence/Medical-R1-Distill-Data},
  urldate = {2026-09-24},
}

@article{gao2026openmopd,
  title={Open-MOPD: Diagnosing and Fixing Capability Imbalance in Multi-Teacher On-Policy Distillation},
  author={Gao, Huan-ang and Chi, Haohan and Yan, Yong and Feng, Shiyuan and Wu, Hanlin and Jiang, Zheng and He, Bingxiang and Ma, Wei-Ying and Zhang, Ya-Qin and Zhou, Hao},
  journal={arXiv preprint arXiv:2608.19098},
  year={2026}
}

@inproceedings{minillm2024,
  title={MiniLLM: Knowledge Distillation of Large Language Models},
  author={Gu, Yuxian and Dong, Li and Wei, Furu and Huang, Minlie},
  booktitle={Proceedings of ICLR},
  year={2024}
}

@misc{hmmt2025feb,
  author = {{Harvard--MIT Mathematics Tournament}},
  title = {Archive of February 2025},
  year = {2025},
  howpublished = {Official tournament archive},
  url = {https://www.hmmt.org/www/archive/282},
  urldate = {2026-09-24},
}

@misc{hmmt2025nov,
  author = {{Harvard--MIT Mathematics Tournament}},
  title = {Archive of November 2025},
  year = {2025},
  howpublished = {Official tournament archive},
  url = {https://www.hmmt.org/www/archive/291},
  urldate = {2026-09-24},
}

@article{he2026simpleopdsimpletokenizeragnosticonpolicy,
  title={SimpleOPD: Simple Tokenizer-Agnostic On-Policy Distillation for Long-Context Reasoning},
  author={Haonan He and Haodi Lei and Yun Luo and Haoran Zhang and Shunkai Zhang and Yizhuo Li and Shengji Tang and Zhilin Wang and Runzhe Zhan and Lei Bai and Ganqu Cui and Fangchen Yu and Yafu Li and Peng Ye and Ning Ding and Yu Cheng},
  journal={arXiv preprint arXiv:2608.14277},
  year={2026}
}

@article{he2025deepmath,
  title={{DeepMath-103K}: A Large-Scale, Challenging, Decontaminated, and Verifiable Mathematical Dataset for Advancing Reasoning},
  author={He, Zhiwei and Liang, Tian and Xu, Jiahao and Liu, Qiuzhi and Chen, Xingyu and Wang, Yue and Song, Linfeng and Yu, Dian and Liang, Zhenwen and Wang, Wenxuan and Zhang, Zhuosheng and Wang, Rui and Tu, Zhaopeng and Mi, Haitao and Yu, Dong},
  journal={arXiv preprint arXiv:2504.11456},
  year={2025}
}

@article{hinton2015KD,
  title = {{Distilling the Knowledge in a Neural Network}},
  author = {Hinton, Geoffrey
    and Vinyals, Oriol
    and Dean, Jeff},
  journal = {arXiv preprint arXiv:1503.02531},
  year = {2015},
  url = {https://arxiv.org/abs/1503.02531},
  eprint = {1503.02531},
  archiveprefix = {arXiv},
}

@article{hou2026uniopd,
  title = {{Uni-OPD: Unifying On-Policy Distillation with a Dual-Perspective Recipe}},
  author = {Wenjin Hou
    and Shangpin Peng
    and Weinong Wang
    and Zheng Ruan
    and Yue Zhang
    and Zhenglin Zhou
    and Mingqi Gao
    and Yifei Chen
    and Kaiqi Wang
    and Hongming Yang
    and Chengquan Zhang
    and Zhuotao Tian
    and Han Hu
    and Yi Yang
    and Fei Wu
    and Hehe Fan},
  journal = {arXiv preprint arXiv:2605.03677},
  year = {2026},
  url = {https://arxiv.org/abs/2605.03677},
  eprint = {2605.03677},
  archiveprefix = {arXiv},
}

@article{huang2025m1,
  title = {{m1: Unleash the Potential of Test-Time Scaling for Medical Reasoning with Large Language Models}},
  author = {Xiaoke Huang
    and Juncheng Wu
    and Hui Liu
    and Xianfeng Tang
    and Yuyin Zhou},
  year = {2025},
  eprint = {2504.00869},
  archiveprefix = {arXiv},
  primaryclass = {cs.CL},
  url = {https://arxiv.org/abs/2504.00869},
  journal = {arXiv preprint arXiv:2504.00869},
}

@misc{intelligentinternet2025iimedical8b,
  author = {{Intelligent Internet}},
  title = {{II-Medical-8B}: Medical Reasoning Model},
  year = {2025},
  howpublished = {Hugging Face model},
  url = {https://huggingface.co/Intelligent-Internet/II-Medical-8B},
  urldate = {2026-09-24},
}

@article{jain2024livecodebench,
  title   = {{LiveCodeBench}: Holistic and Contamination Free Evaluation of Large Language Models for Code},
  author  = {Jain, Naman and Han, King and Gu, Alex and Li, Wen-Ding and Yan, Fanjia and Zhang, Tianjun and Wang, Sida and Solar-Lezama, Armando and Sen, Koushik and Stoica, Ion},
  journal = {arXiv preprint arXiv:2403.07974},
  year    = {2024}
}

@article{jin2020medqa,
  title   = {What Disease Does This Patient Have? A Large-Scale Open Domain Question Answering Dataset from Medical Exams},
  author  = {Jin, Di and Pan, Eileen and Oufattole, Nassim and Weng, Wei-Hung and Fang, Hanyi and Szolovits, Peter},
  journal = {arXiv preprint arXiv:2009.13081},
  year    = {2020}
}

@inproceedings{kim2016eqkd,
  title={Sequence-Level Knowledge Distillation},
  author={Kim, Yoon and Rush, Alexander M.},
  editor={Su, Jian and Duh, Kevin and Carreras, Xavier},
  booktitle={Proceedings of the 2016 Conference on Empirical Methods in Natural Language Processing},
  month=nov,
  year={2016},
  address={Austin, Texas},
  publisher={Association for Computational Linguistics},
  url={https://aclanthology.org/D16-1139/},
  doi={10.18653/v1/D16-1139},
  pages={1317--1327}
}

@InProceedings{distillm2025,
  title={{D}isti{LLM}-2: A Contrastive Approach Boosts the Distillation of {LLM}s},
  author={Ko, Jongwoo and Chen, Tianyi and Kim, Sungnyun and Ding, Tianyu and Liang, Luming and Zharkov, Ilya and Yun, Se-Young},
  booktitle={Proceedings of the 42nd International Conference on Machine Learning},
  pages={31044--31062},
  year={2025},
  editor={Singh, Aarti and Fazel, Maryam and Hsu, Daniel and Lacoste-Julien, Simon and Berkenkamp, Felix and Maharaj, Tegan and Wagstaff, Kiri and Zhu, Jerry},
  volume={267},
  series={Proceedings of Machine Learning Research},
  month={13--19 Jul},
  publisher={PMLR},
  url={https://proceedings.mlr.press/v267/ko25a.html}
}

@article{li2026filter,
  title={Filter, Then Reweight: Rethinking Optimization Granularity in On-Policy Distillation},
  author={Yuying Li and Leqi Zheng and Yongzi Yu and Wenrui Zhou and Xuchang Zhong and Xing Hu and Jing Jin and Hangjie Yuan and Tao Feng},
  journal={arXiv preprint arXiv:2606.02684},
  year={2026}
}

@article{li2026coinsidesdualnature,
  title={Every Coin Has Two Sides: On the Dual Nature of Generalization in On-Policy Distillation of Large Language Models},
  author={Zhaoyi Li and Deyang Kong and Yuan Wei and Evan Yang and Ranran Shen and Mahardika Krisna Ihsani and Ming Yang and Wei Zhang and Chuan Hao and Jian Yang and Ran Tao and Bryan Dai and Shikun Zhang and Wei Ye and Ying Wei and Defu Lian},
  journal={arXiv preprint arXiv:2608.16647},
  year={2026}
}

@article{liang2025modomodomultidomaindatamixtures,
  title={MoDoMoDo: Multi-Domain Data Mixtures for Multimodal LLM Reinforcement Learning},
  author={Yiqing Liang and Jielin Qiu and Wenhao Ding and Zuxin Liu and James Tompkin and Mengdi Xu and Mengzhou Xia and Zhengzhong Tu and Laixi Shi and Jiacheng Zhu},
  journal={arXiv preprint arXiv:2505.24871},
  year={2025}
}

@article{lin2026opdwithverifiablereward,
  title={On-policy Distillation with Verifiable Reward},
  author={Wenze Lin and Jiale Zhao and Xitai Jiang and Songde Rao and Yining Li and Shenzhi Wang and Bingxiang He and Gao Huang},
  journal={arXiv preprint arXiv:2608.24696},
  year={2026}
}

@InProceedings{NeurIPS2021-Liu,
  author={Bo Liu and Xingchao Liu and Xiaojie Jin and Peter Stone and Qiang Liu},
  title={Conflict-Averse Gradient Descent for Multi-task learning},
  booktitle={Conference on Neural Information Processing Systems (NeurIPS), 2021},
  location={Virtual Only},
  month={December},
  year={2021}
}

@inproceedings{liu2023evalplus,
  title = {{Is Your Code Generated by ChatGPT Really Correct? Rigorous Evaluation of Large Language Models for Code Generation}},
  author = {Liu, Jiawei
    and Xia, Chunqiu Steven
    and Wang, Yuyao
    and Zhang, Lingming},
  booktitle = {Thirty-Seventh Conference on Neural Information Processing Systems},
  year = {2023},
  url = {https://proceedings.neurips.cc/paper_files/paper/2023/hash/43e9d647ccd3e4b7b5baab53f0368686-Abstract.html},
  eprint = {2305.01210},
  archiveprefix = {arXiv},
}

@article{mapteam2025supergpqa,
  title = {{SuperGPQA: Scaling LLM Evaluation across 285 Graduate Disciplines}},
  author = {{M-A-P Team}
    and Xinrun Du
    and Yifan Yao
    and Kaijing Ma
    and Bingli Wang
    and Tianyu Zheng
    and others},
  year = {2025},
  eprint = {2502.14739},
  archiveprefix = {arXiv},
  primaryclass = {cs.CL},
  url = {https://arxiv.org/abs/2502.14739},
  journal = {arXiv preprint arXiv:2502.14739},
}

@article{ma2026mopdmultiteacheronpolicydistillation,
  title={MOPD: Multi-Teacher On-Policy Distillation for Capability Integration in LLM Post-Training},
  author={Wenhan Ma and Jianyu Wei and Liang Zhao and Hailin Zhang and Bangjun Xiao and Lei Li and Qibin Yang and Bofei Gao and Yudong Wang and Rang Li and Jinhao Dong and Zhifang Sui and Fuli Luo},
  journal={arXiv preprint arXiv:2606.30406},
  year={2026}
}

@misc{maa2024aime,
  author = {{Mathematical Association of America}},
  shortauthor = {{MAA}},
  title = {{2024 AIME I Problems}},
  year = {2024},
  howpublished = {Problem sets archived by Art of Problem Solving},
  url = {https://artofproblemsolving.com/wiki/index.php?title=2024_AIME_I_Problems},
}

@misc{maa2025aime,
  author = {{Mathematical Association of America}},
  shortauthor = {{MAA}},
  title = {{2025 AIME I Problems}},
  year = {2025},
  howpublished = {Problem sets archived by Art of Problem Solving},
  url = {https://artofproblemsolving.com/wiki/index.php?title=2025_AIME_I_Problems},
}

@techreport{GPT4,
  author = {{OpenAI}},
  title = {{GPT-4 Technical Report}},
  year = {2023},
  url = {https://arxiv.org/abs/2303.08774},
  eprint = {2303.08774},
  archiveprefix = {arXiv},
  institution = {OpenAI},
  number = {arXiv:2303.08774},
}

@inproceedings{pal2022medmcqa,
  title     = {{MedMCQA}: A Large-Scale Multi-Subject Multi-Choice Dataset for Medical Domain Question Answering},
  author    = {Pal, Ankit and Umapathi, Logesh Kumar and Sankarasubbu, Malaikannan},
  booktitle = {Proceedings of the Conference on Health, Inference, and Learning},
  pages     = {248--260},
  year      = {2022},
  volume    = {174},
  series    = {Proceedings of Machine Learning Research},
  publisher = {PMLR},
  url       = {https://proceedings.mlr.press/v174/pal22a.html}
}

@article{qwen2025qwen3,
  title = {{Qwen3 Technical Report}},
  author = {An Yang
    and Anfeng Li
    and Baosong Yang
    and Beichen Zhang
    and Binyuan Hui
    and Bo Zheng
    and others},
  journal = {arXiv preprint arXiv:2505.09388},
  year = {2025},
  url = {https://arxiv.org/abs/2505.09388},
  eprint = {2505.09388},
  archiveprefix = {arXiv},
}

@article{shao2026tokenlevelanalys,
  title = {{A Token-Level Analysis of Sampled-Token Reverse-KL On-Policy Distillation}},
  author = {Bing Shao
    and Jiazheng Zhang
    and Long Ma
    and Yujiong Shen
    and Senjie Jin
    and Xin Guo
    and Yuming Yang
    and Mingxu Chai
    and Zhiheng Xi
    and Boyang Liu
    and Junlin Shang
    and Tao Gui
    and Qi Zhang
    and Xuanjing Huang},
  year = {2026},
  eprint = {2608.25643},
  archiveprefix = {arXiv},
  primaryclass = {cs.LG},
  url = {https://arxiv.org/abs/2608.25643},
  journal = {arXiv preprint arXiv:2608.25643},
}

@article{sun2026steplevelonpolicydistillationinterpolating,
  title={Step-Level On-Policy Distillation: Interpolating Between On-Policy Distillation and Supervised Fine-Tuning},
  author={Changhui Sun and Lanbo Liu and Hang Lei and Tong Ling and Jiahang Xie and Zhiyong Zheng and Yujia Wang and Hao Liu and Feng Xiao and Lu Liu and Yanlong Du and Zifeng Cheng and Ziwei Jiang and Qing Gu},
  journal={arXiv preprint arXiv:2608.16333},
  year={2026}
}

@article{sun2026d3mopdadaptivedynamicdomain,
  title = {{D$^3$-MOPD: Dynamic Domain ScheDuling for Efficient Multi-Teacher Distillation}},
  author = {Zechen Sun
    and Zhiwei Zhang
    and Fei Zhao
    and Juntao Li
    and Mu Chuan
    and Huayu Deng
    and Guojian Zhan
    and Wenliang Chen
    and Yao Hu
    and Min Zhang},
  journal = {arXiv preprint arXiv:2608.24987},
  year = {2026},
  url = {https://arxiv.org/abs/2608.24987v2},
  eprint = {2608.24987},
  archiveprefix = {arXiv},
}

@article{wu2025medreason,
  title         = {{MedReason}: Eliciting Factual Medical Reasoning Steps in {LLMs} via Knowledge Graphs},
  author        = {Juncheng Wu and Wenlong Deng and Xingxuan Li and Sheng Liu and Taomian Mi and Yifan Peng and Ziyang Xu and Yi Liu and Hyunjin Cho and Chang-In Choi and Yihan Cao and Hui Ren and Xiang Li and Xiaoxiao Li and Yuyin Zhou},
  journal={arXiv preprint arXiv:2504.00993},
  year          = {2025}
}

@article{xiao2026mimov2flashtechnicalreport,
  title = {{MiMo-V2-Flash Technical Report}},
  author = {{Xiaomi LLM-Core Team}
    and Bangjun Xiao
    and Bingquan Xia
    and Bo Yang
    and Bofei Gao
    and Bowen Shen
    and others},
  journal = {arXiv preprint arXiv:2601.02780},
  year = {2026},
  url = {https://arxiv.org/abs/2601.02780},
  eprint = {2601.02780},
  archiveprefix = {arXiv},
}

@article{xie2023doremi,
  title={DoReMi: Optimizing Data Mixtures Speeds Up Language Model Pretraining},
  author={Xie, Sang Michael and Pham, Hieu and Dong, Xuanyi and Du, Nan and Liu, Hanxiao and Lu, Yifeng and Liang, Percy and Le, Quoc V. and Ma, Tengyu and Yu, Adams Wei},
  journal={Advances in Neural Information Processing Systems},
  volume={36},
  pages={69798--69818},
  year={2023}
}

@article{yang2026filteropd,
  title={Beyond Imitation: Filtering On-Policy Distillation by Reasoning Progress},
  author={Chen Yang and Haiyuan Wan and Rengrong Xiong and Yize Chen and Danny H. K. Tsang},
  journal={arXiv preprint arXiv:2608.19408},
  year={2026}
}

@article{yang2026gopd,
  title   = {Learning Beyond Teacher: Generalized On-Policy Distillation with Reward Extrapolation},
  author  = {Yang, Wenkai and Liu, Weijie and Xie, Ruobing and Yang, Kai and Yang, Saiyong and Lin, Yankai},
  journal = {arXiv preprint arXiv:2602.12125},
  year    = {2026},
  url     = {https://arxiv.org/abs/2602.12125}
}

@article{Nemotron_Cascade_2,
  title = {{Nemotron-Cascade 2: Post-Training LLMs with Cascade RL and Multi-Domain On-Policy Distillation}},
  author = {Zhuolin Yang
    and Zihan Liu
    and Yang Chen
    and Wenliang Dai
    and Boxin Wang
    and Sheng-Chieh Lin
    and Chankyu Lee
    and Yangyi Chen
    and Dongfu Jiang
    and Jiafan He
    and Renjie Pi
    and Grace Lam
    and Nayeon Lee
    and Alexander Bukharin
    and Mohammad Shoeybi
    and Bryan Catanzaro
    and Wei Ping},
  year = {2026},
  eprint = {2603.19220},
  archiveprefix = {arXiv},
  primaryclass = {cs.CL},
  url = {https://arxiv.org/abs/2603.19220},
  journal = {arXiv preprint arXiv:2603.19220},
}

@article{yin2026hopdconfidenceawareheterogeneous,
  title={H-OPD: Confidence Aware Heterogeneous Multi-Teacher Multimodal On-policy Distillation},
  author={Qixiang Yin and Huanjin Yao and Yuchen Cai and Jianghao Chen and Ziyi Wang and Min Yang and Fei Su and Zhicheng Zhao},
  journal={arXiv preprint arXiv:2607.02592},
  year={2026}
}

@article{yu2020gradient,
  title={Gradient Surgery for Multi-Task Learning},
  author={Yu, Tianhe and Kumar, Saurabh and Gupta, Abhishek and Levine, Sergey and Hausman, Karol and Finn, Chelsea},
  journal={Advances in Neural Information Processing Systems},
  volume={33},
  pages={5824--5836},
  year={2020}
}

@article{zhang2026teacherlikelihoodgroupcalibratedonpolicy,
  title={Beyond Teacher Likelihood: Group-Calibrated On-Policy Distillation for Long-Context Reasoning},
  author={Zhu Zhang and Jixun Wang and Xiaoang Xu and Xiaorong Wang and Zihan Zhou and Zhiyuan Wang and Shuo Wang and Chaojun Xiao and Yuezhi Zhou},
  journal={arXiv preprint arXiv:2608.19181},
  year={2026}
}

@inproceedings{zuo2025medxpertqa,
  title     = {{MedXpertQA}: Benchmarking Expert-Level Medical Reasoning and Understanding},
  author    = {Zuo, Yuxin and Qu, Shang and Li, Yifei and Chen, Zhang-Ren and Zhu, Xuekai and Hua, Ermo and Zhang, Kaiyan and Ding, Ning and Zhou, Bowen},
  booktitle = {Proceedings of the 42nd International Conference on Machine Learning},
  pages     = {80961--80990},
  year      = {2025},
  volume    = {267},
  series    = {Proceedings of Machine Learning Research},
  publisher = {PMLR},
  url       = {https://proceedings.mlr.press/v267/zuo25a.html}
}
\bibliographystyle{iclr2027_conference}

\clearpage
\appendix

\section*{Appendix Contents}

\begingroup
\hypersetup{linkcolor=black}
\setlength{\parindent}{0pt}
\setlength{\parskip}{3pt}
\newcommand{\appcontentsline}[2]{%
    \noindent\hspace*{#1}%
    \hyperref[#2]{\makebox[2.5em][l]{\ref*{#2}}\nameref*{#2}}%
    \nobreak\dotfill\hyperref[#2]{\pageref*{#2}}\par}
\medskip
\hfill\textbf{Page}\par
{\bfseries\appcontentsline{0pt}{app:method-details}}
\appcontentsline{1.5em}{app:distillation-objective}
\appcontentsline{1.5em}{app:projection-derivation}
\appcontentsline{1.5em}{app:training-algorithm}
\medskip
{\bfseries\appcontentsline{0pt}{app:experimental-setup}}
\appcontentsline{1.5em}{app:models}
\appcontentsline{1.5em}{app:training-data}
\appcontentsline{1.5em}{app:hyperparameters}
\appcontentsline{1.5em}{app:baseline-details}
\appcontentsline{1.5em}{app:evaluation-protocols}
\medskip
{\bfseries\appcontentsline{0pt}{app:additional-results}}
\appcontentsline{1.5em}{app:benchmark-details}
\appcontentsline{1.5em}{app:low-conflict-results}
\appcontentsline{1.5em}{app:extended-token-visualizations}
\appcontentsline{1.5em}{app:training-dynamics}
\medskip
{\bfseries\appcontentsline{0pt}{app:limitations}}
\endgroup
\clearpage

\section{Additional Method Details}
\label{app:method-details}

\subsection{OPD Derivation and Training Loss}
\label{app:distillation-objective}

\paragraph{From sequence-level reverse KL to token-level supervision.}
We expand the OPD reformulation in Section~\ref{sec:method:setup}
before specifying its sampled-token implementation. For a prompt
\(x\sim\mathcal D\), let \(y=(y_1,\ldots,y_T)\sim\pi_\theta(\cdot\mid x)\)
and \(h_t=(x,y_{<t})\). Write
\(p_t=\pi_\theta(\cdot\mid h_t)\) and
\(q_t=\pi_T(\cdot\mid h_t)\) for the student and teacher distributions
on the same prefix. With \(\mathbb E_\theta\) denoting expectation over
these prompts and student-generated responses, autoregressive
factorization and the KL chain rule give
\begin{equation}
\begin{aligned}
    \mathcal L_{\mathrm{OPD}}(\theta)
    &=\mathbb E_{x\sim\mathcal D}
      D_{\mathrm{KL}}\!\left(\pi_\theta(\cdot\mid x)\,\|\,\pi_T(\cdot\mid x)\right)\\
    &=\mathbb E_\theta\!\left[\log\frac{\pi_\theta(y\mid x)}{\pi_T(y\mid x)}\right]\\
    &=\mathbb E_\theta\!\left[\sum_{t=1}^{T}
      \log\frac{p_t(y_t)}{q_t(y_t)}\right]
     =\mathbb E_\theta\!\left[\sum_{t=1}^{T}D_{\mathrm{KL}}(p_t\,\|\,q_t)\right].
\end{aligned}
    \label{eq:app-opd-chain}
\end{equation}
The expectations assume finite divergences and common policy support.
Thus, the token-level teacher distributions are evaluated on prefixes
visited by the student.

\paragraph{Equivalent relative-reward objective.}
Introducing a frozen reference \(\pi_{\mathrm{ref}}\) and adding and
subtracting its log-probability exposes the teacher's relative reward
\citep{yang2026gopd}:
\begin{equation}
\begin{aligned}
    -\mathcal L_{\mathrm{OPD}}(\theta)
    &=\mathbb E_\theta\!\left[\log\pi_T(y\mid x)-\log\pi_\theta(y\mid x)\right]\\
    &=\mathbb E_\theta\!\left[
       \bigl(\log\pi_T(y\mid x)-\log\pi_{\mathrm{ref}}(y\mid x)\bigr)
       -\bigl(\log\pi_\theta(y\mid x)-\log\pi_{\mathrm{ref}}(y\mid x)\bigr)
      \right]\\
    &=\mathbb E_\theta\!\left[\log\frac{\pi_T(y\mid x)}{\pi_{\mathrm{ref}}(y\mid x)}\right]
      -\mathbb E_{x\sim\mathcal D}
       D_{\mathrm{KL}}\!\left(\pi_\theta(\cdot\mid x)\,\|\,\pi_{\mathrm{ref}}(\cdot\mid x)\right)
      \equiv\mathcal J_1(\theta).
\end{aligned}
    \label{eq:app-opd-reference}
\end{equation}
Consequently, \(\arg\min_\theta\mathcal L_{\mathrm{OPD}}(\theta)
=\arg\max_\theta\mathcal J_1(\theta)\): the reward-maximization
objective is the negative of the reverse-KL loss. Scaling only the
relative-reward term by \(\lambda>0\) gives the objective in
Eq.~(\ref{eq:lambda-opd}), whose sampled-response form is
\begin{equation}
\begin{aligned}
    \mathcal J_\lambda(\theta)
    &=\mathbb E_\theta\!\left[
      \lambda\log\frac{\pi_T(y\mid x)}{\pi_{\mathrm{ref}}(y\mid x)}
      -\log\frac{\pi_\theta(y\mid x)}{\pi_{\mathrm{ref}}(y\mid x)}\right]\\
    &=\mathbb E_\theta\!\left[\sum_{t=1}^{T}
      \left\{\lambda\bigl(\ell_t^T-\ell_t^{\mathrm{ref}}\bigr)
      -\bigl(\ell_t^\theta-\ell_t^{\mathrm{ref}}\bigr)\right\}\right],
\end{aligned}
    \label{eq:app-scaled-opd}
\end{equation}
where \(\ell_t^T\), \(\ell_t^\theta\), and \(\ell_t^{\mathrm{ref}}\)
are the teacher, student, and reference log-probabilities of the sampled
token. At \(\lambda=1\), the reference terms cancel and recover OPD;
\(0<\lambda<1\) attenuates the relative reward, whereas \(\lambda>1\)
amplifies it (ExOPD). Domain routing applies this construction to each
teacher--domain pair in Eq.~(\ref{eq:mopd}).

\paragraph{Credit assignment with multiple teachers.}
For the \(K>2\) extension of Section~\ref{sec:method:token},
let \(\mathcal P_d=\{j\mid j\neq d\}\) be the peer teachers for target
domain \(d\). Here we use explicit teacher indices: for each peer \(j\),
the main-text roles are \(T_{\mathrm r}=T_d\) and \(T_{\mathrm a}=T_j\),
with \(\ell_t^{\mathrm r}=\ell_t^d\) and \(\ell_t^{\mathrm a}=\ell_t^j\).
All peers score the same trajectory as the routed teacher,
and their token agreements are averaged:
\begin{equation}
    \bar s_{d,t}
    =\frac{1}{K-1}\sum_{j\in\mathcal P_d}
      \tanh\!\left(\frac{\ell_t^d-\ell_t^{\mathrm{ref}}}{\tau}\right)
      \tanh\!\left(\frac{\ell_t^j-\ell_t^{\mathrm{ref}}}{\tau}\right).
    \label{eq:multi-peer-agreement}
\end{equation}
We obtain \(\lambda_{d,t}\) by applying Eq.~(\ref{eq:token-wise-lambda})
with \(s_t=\bar s_{d,t}\). For \(K=2\), this recovers
Eqs.~(\ref{eq:teacher-reward-agreement})--(\ref{eq:token-wise-lambda}).
For \(K>2\), the \(K-1\) peers reinforce or attenuate the routed signal
according to their mean evidence; opposing assessments can cancel,
recovering standard OPD when \(\bar s_{d,t}=0\).

\paragraph{Sampled-token learning signal.}
The implementation replaces the fixed reward scale in
Eq.~(\ref{eq:app-scaled-opd}) with the agreement coefficient from
Section~\ref{sec:method:token}, using the sampled-token \(k_1\) log-ratio estimator
and a PPO-style policy surrogate. Let \(\pi_{\mathrm{old}}\) denote the student policy that generates a
stored response, and let \(\ell_{i,t}^{\theta}=\log\pi_\theta(y_{i,t}\mid h_{i,t})\)
and \(\ell_{i,t}^{\mathrm{old}}=\log\pi_{\mathrm{old}}(y_{i,t}\mid h_{i,t})\).
For a response from domain \(d\), the routed teacher and shared frozen reference
define \(\Delta^d_{i,t}=\ell^d_{i,t}-\ell^{\mathrm{ref}}_{i,t}\).
The advantage supplied to the policy loss is
\begin{equation}
    A^{\mathrm{agr}}_{d,i,t}
    =\operatorname{sg}\!\left[
       \operatorname{clip}\!\left(
       \lambda_{d,i,t}\Delta^d_{i,t}
       -\bigl(\ell^\theta_{i,t}-\ell^{\mathrm{ref}}_{i,t}\bigr),-10,10
       \right)
     \right],
    \label{eq:app-advantage}
\end{equation}
where \(\operatorname{sg}\) denotes stop-gradient. The entire recomputed advantage
is clipped to \([-10,10]\) after agreement modulation and detached before entering
the policy surrogate, including its student log-probability term.
Teacher and reference parameters are frozen throughout training.
When \(\lambda_{d,i,t}=1\), Eq.~(\ref{eq:app-advantage}) reduces to the plain
OPD learning signal
\(\operatorname{sg}[\operatorname{clip}(\ell^d_{i,t}-\ell^\theta_{i,t},-10,10)]\).
Thus, the coefficient rescales the teacher's reward relative to the reference;
it does not multiply the entire OPD loss.

\paragraph{Clipped policy surrogate and masking.}
The importance ratio and per-token loss are
\begin{align}
    r_{i,t}(\theta)
      &=\exp\!\left(\ell^\theta_{i,t}-\ell^{\mathrm{old}}_{i,t}\right),
      \label{eq:app-ratio}\\
    \ell^{\mathrm{agr}}_{d,i,t}(\theta)
      &=-\min\!\left\{
        r_{i,t}(\theta)A^{\mathrm{agr}}_{d,i,t},\,
        \operatorname{clip}\!\left(r_{i,t}(\theta),1-\epsilon_{\mathrm{PPO}},
        1+\epsilon_{\mathrm{PPO}}\right)A^{\mathrm{agr}}_{d,i,t}
      \right\}.
      \label{eq:app-clipped-loss}
\end{align}
Let \(M_{i,t}\) indicate a valid response token and
\(T_i=\sum_t M_{i,t}\). The target-domain loss is
\begin{equation}
    \mathcal L_d^{\mathrm{agr}}
      =\frac{1}{|\mathcal B_d|}\sum_{i\in\mathcal B_d}
       \frac{\sum_t M_{i,t}\ell^{\mathrm{agr}}_{d,i,t}}{T_i}.
    \label{eq:app-masked-reduction}
\end{equation}
Prompt and padding positions do not contribute to the loss. The recorded training
configuration uses \(\epsilon_{\mathrm{PPO}}=0.2\), without an outcome-based task
reward or a separate KL reward added by the trainer.
The single-turn rollout mask retains the returned generated tokens, including
EOS when returned, and excludes padding. A response reaching the length cap
retains supervision on its available prefix. Fully masked responses contribute zero.
The agreement coefficient \(\lambda_{d,i,t}\) is bounded by \([0.5,1.5]\).

\paragraph{Auxiliary supervision.}
For each peer teacher \(j\neq d\), we reuse the same stored domain-\(d\) responses
and replace the advantage by
\(A^{\mathrm{OPD}}_{d\rightarrow j,i,t}
=\operatorname{sg}[\operatorname{clip}(\ell^j_{i,t}-\ell^\theta_{i,t},-10,10)]\).
The auxiliary loss uses the same clipped surrogate, with agreement
modulation disabled and the same advantage clipping range. For the selected
auxiliary teacher \(T_{\mathrm a}=T_j\), we write the main-text loss
\(\mathcal L'_{d\rightarrow\mathrm a}\) as \(\mathcal L'_{d\rightarrow j}\).
The full method and
all ablations retain this range, including when credit assignment is disabled.
Advantage clipping is
separate from the importance-ratio clipping in Eq.~(\ref{eq:app-clipped-loss}).

\subsection{Derivation of Norm-Constrained Projection}
\label{app:projection-derivation}

For nonzero \(g\) and \(g'\), let
\(c=\langle g,g'\rangle/(\|g\|_2\|g'\|_2)\).
When \(c<0\), every candidate in \(\mathcal S(g,g')\) can be written as
\begin{equation}
    v(\alpha)=g-\alpha\frac{\langle g,g'\rangle}{\|g'\|_2^2}g',
    \qquad 0\leq\alpha\leq1.
    \label{eq:app-partial-projection}
\end{equation}
The objective in Eq.~(\ref{eq:projection-optimization}) is
\(-(1-\alpha)\langle g,g'\rangle\), which strictly decreases with
\(\alpha\). The optimum therefore uses the largest feasible coefficient.
Expanding the squared norm gives
\begin{equation}
    \frac{\|v(\alpha)\|_2^2}{\|g\|_2^2}
    =1-(2\alpha-\alpha^2)c^2.
    \label{eq:app-retention-expansion}
\end{equation}
This decreases monotonically on \([0,1]\). Full projection is feasible
if \(\sqrt{1-c^2}\geq m\); otherwise, the optimum satisfies
\(1-(2\alpha-\alpha^2)c^2=m^2\). Including the no-conflict case gives
\begin{equation}
    \alpha^\star=
    \begin{cases}
        0, & c\geq0,\\
        1, & c<0\ \text{and}\ \sqrt{1-c^2}\geq m,\\
        1-\sqrt{1-\dfrac{1-m^2}{c^2}},
           & c<0\ \text{and}\ \sqrt{1-c^2}<m.
    \end{cases}
    \label{eq:app-alpha-derivation}
\end{equation}
Substitution into Eq.~(\ref{eq:v2-projection}) recovers the unique
solution of Eq.~(\ref{eq:projection-optimization}) for a conflicting pair.

The norm constraint trades complete conflict removal for retention of the
target-domain signal. In particular,
\begin{equation}
    \langle\widetilde g,g'\rangle
    =(1-\alpha^\star)\langle g,g'\rangle,
    \label{eq:app-residual-conflict}
\end{equation}
which remains negative when \(c<0\) and \(\alpha^\star<1\). These statements concern
the projected tensor before domain aggregation, global clipping, and AdamW
preconditioning. They do not guarantee a non-increasing auxiliary loss after
the final optimizer step, nor a lower bound on the norm of the merged update.

\paragraph{Multi-teacher projection and shared norm retention.}
For \(K>2\), each peer \(j\in\mathcal P_d\) induces a same-trajectory
auxiliary gradient \(g'_{d\rightarrow j}\), computed as in
Eq.~(\ref{eq:v2-auxiliary-gradient}). For each parameter tensor, we select
the remaining valid peer with the most negative cosine, project against it,
and recompute the cosines; processing stops when no valid negative cosine
remains. These projections use the norm-constrained form of
Eq.~(\ref{eq:v2-projection}) with a shared floor relative to the original
target tensor:
\begin{equation}
    \|g_d^{(q)}\|_2\geq m\|g_d^{(0)}\|_2,
    \qquad q=1,\ldots,K-1,
    \label{eq:app-shared-retention}
\end{equation}
where \(g_d^{(0)}\) is the original target tensor and \(g_d^{(q)}\)
is the tensor after \(q\) projections. Once processing is complete,
the resulting tensors form \(\widetilde g_d\). The floor is measured
against the original tensor throughout the sequence. Applying a fresh local
retention ratio \(m\) independently at every projection would provide only
a product bound and would not implement this shared constraint. In addition,
a later projection can change alignment with a previously processed peer;
sequential projection alone does not establish simultaneous non-conflict
with every peer.

For each conflicting peer, the implementation computes the coefficient from
the remaining squared-norm budget. Write \(a=\|g_d^{(q-1)}\|_2^2\),
\(b=m^2\|g_d^{(0)}\|_2^2\), and
\(u=\langle g_d^{(q-1)},g'\rangle^2/\|g'\|_2^2\).
It uses \(\alpha=1\) if \(a-u\geq b\); otherwise,
\(\alpha=1-\sqrt{1-\operatorname{clip}((a-b)/u,0,1)}\).
Thus, a tensor that has exhausted its shared budget receives no further
projection. These comparisons use numerical tolerance \(10^{-12}\).

\subsection{Complete Training Procedure}
\label{app:training-algorithm}

Algorithm~\ref{alg:roadmopd-step} combines teacher scoring, token modulation,
and all-peer gradient projection within one optimizer step, using the
domain-mean gradient convention of the main text. All target and auxiliary
gradients are evaluated at the same student parameters on stored
trajectories; only projected target gradients enter the update. Batch
normalization is detailed below, and archived projection configurations are
specified in Appendix~\ref{app:hyperparameters}.

\begin{algorithm}[!t]
\caption{One optimizer step of \method}
\label{alg:roadmopd-step}
\small
\begin{algorithmic}[1]
    \Require Student parameters $\theta$, optimizer state $\mathcal S$, rollout policy $\pi_{\mathrm{old}}$
    \Require Frozen teachers $\{\pi_{T_d}\}_{d=1}^{K}$, frozen reference $\pi_{\mathrm{ref}}$, labelled prompts $\mathcal X$
    \Require Modulation parameters $(\tau,\rho_{\mathrm g},\lambda_{\min},\lambda_{\max})$, retention $m$, clip threshold $\gamma$
    \Ensure Updated student parameters $\theta$ and optimizer state $\mathcal S$
    \State $\mathcal B,\ell^{\mathrm{old}},M \gets \Call{Rollout}{\pi_{\mathrm{old}},\mathcal X}$
    \State $\ell^{1:K},\ell^{\mathrm{ref}} \gets \Call{ScoreSameTrajectories}{\{\pi_{T_d}\}_{d=1}^{K},\pi_{\mathrm{ref}},\mathcal B}$
    \State $g_{\mathrm{update}}\gets 0$
    \For{$d=1,\ldots,K$}
        \State $\mathcal B_d\gets\{i\in\mathcal B:\operatorname{domain}(i)=d\}$; $\mathcal P_d\gets\{1,\ldots,K\}\setminus\{d\}$
        \State $\ell^\theta\gets\Call{StudentLogProbs}{\pi_\theta,\mathcal B_d}$; $r\gets\exp(\ell^\theta-\ell^{\mathrm{old}})$
        \State $\bar s_{d,i,t}\gets\frac{1}{K-1}\sum_{j\in\mathcal P_d}
        \tanh\!\bigl((\ell^d_{i,t}-\ell^{\mathrm{ref}}_{i,t})/\tau\bigr)
        \tanh\!\bigl((\ell^j_{i,t}-\ell^{\mathrm{ref}}_{i,t})/\tau\bigr)$
        \State $\lambda_{d,i,t}\gets\operatorname{clip}(1+\rho_{\mathrm g}\bar s_{d,i,t},\lambda_{\min},\lambda_{\max})$
        \State $A^{\mathrm{agr}}_{d,i,t}\gets\operatorname{sg}\!\left[
        \operatorname{clip}\!\left(
        \lambda_{d,i,t}(\ell^d_{i,t}-\ell^{\mathrm{ref}}_{i,t})
        -(\ell^\theta_{i,t}-\ell^{\mathrm{ref}}_{i,t}),-10,10\right)\right]$
        \State $\mathcal L_d^{\mathrm{agr}}\gets\Call{MaskedLoss}{r,A_d^{\mathrm{agr}},M_d}$
        \State $g_d\gets\nabla_\theta\mathcal L_d^{\mathrm{agr}}$
        \For{$j\in\mathcal P_d$}
            \State $A^{\mathrm{OPD}}_{d\rightarrow j,i,t}\gets\operatorname{sg}[\operatorname{clip}(\ell^j_{i,t}-\ell^\theta_{i,t},-10,10)]$
            \State $\mathcal L'_{d\rightarrow j}\gets\Call{MaskedLoss}{r,A^{\mathrm{OPD}}_{d\rightarrow j},M_d}$
            \State $g'_{d\rightarrow j}\gets\nabla_\theta\mathcal L'_{d\rightarrow j}$
        \EndFor
        \If{$K=2$}
            \State $\widetilde g_d\gets\Call{TensorwiseProject}{g_d,g'_{d\rightarrow j},m}$, where $\mathcal P_d=\{j\}$
        \Else
            \State $\widetilde g_d\gets\Call{ProjectPeersSharedFloor}{g_d,\{g'_{d\rightarrow j}\}_{j\in\mathcal P_d},m}$
        \EndIf
        \State $g_{\mathrm{update}}\gets g_{\mathrm{update}}+\frac{1}{K}\widetilde g_d$
    \EndFor
    \State $g_{\mathrm{update}}\gets\Call{GlobalNormClip}{g_{\mathrm{update}},\gamma}$
    \State $(\theta,\mathcal S)\gets\Call{AdamWStep}{\theta,\mathcal S,g_{\mathrm{update}}}$
    \State \Return $\theta,\mathcal S$
\end{algorithmic}
\end{algorithm}

Token operations are elementwise over the valid responses in $\mathcal B_d$;
stored scores and masks are restricted to that domain when used.
\textsc{MaskedLoss} computes the loss in
Eqs.~(\ref{eq:app-clipped-loss}) and (\ref{eq:app-masked-reduction})
using valid response tokens.
\textsc{TensorwiseProject} applies Eq.~(\ref{eq:v2-projection}) separately
to each parameter tensor. \textsc{ProjectPeersSharedFloor} denotes the
sequential multi-peer procedure detailed in Appendix~\ref{app:projection-derivation},
subject to the shared norm floor in Eq.~(\ref{eq:app-shared-retention}). Auxiliary gradients are
discarded after projection and never merged into $g_{\mathrm{update}}$.

\paragraph{Domain means and global-batch normalization.}
Equation~(\ref{eq:app-masked-reduction}) and the algorithm define each
gradient from a domain mean, and the algorithm averages the projected
gradients across domains as in Eq.~(\ref{eq:merged-update}).
In the implementation path supplied with
full-batch normalization metadata, masking or slicing to domain \(d\)
retains the denominator \(|\mathcal B|\). After microbatch accumulation and
data-parallel averaging, its loss contribution is therefore
\(\mathcal L_d^{\mathrm{batch}}=w_d\mathcal L_d^{\mathrm{agr}}\), where
\(w_d=|\mathcal B_d|/|\mathcal B|\); auxiliary losses on the same responses
receive the same factor. Scaling both gradients by \(w_d>0\) preserves
the projection cosines and retention coefficients in exact arithmetic.
To express this implementation using the algorithm's domain-mean gradients,
replace its accumulation by
\(g_{\mathrm{update}}\gets g_{\mathrm{update}}+w_d\widetilde g_d\).
Balanced batches give \(w_d=1/K\), recovering the equal-weight average in
Eq.~(\ref{eq:merged-update}); shuffled batches use their observed domain
fractions. For imbalanced batches, these weights differ from \(1/K\)
before global clipping and AdamW, so the two weighting conventions are
not assumed to produce identical optimizer steps.

\paragraph{Projection implementation.}
The paired-domain implementation projects each trainable FSDP parameter
tensor separately. Gradient snapshots are stored in BF16, while dot
products and squared norms are computed in FP64 and reduced across the
data-parallel group. Shards of the same tensor therefore use a common
cosine and projection coefficient. Tensors with negligible gradient norms
are left unchanged; the numerical stability constant is \(10^{-12}\).
With multiple peers, each tensor selects the most negative cosine among its
unprocessed peers and recomputes these cosines after each projection. Ties
are broken by the configured peer order. Peers with negligible norms are
skipped, and processing stops when no valid negative cosine remains. A
selected peer is removed from the candidate set, even if the norm budget
forces its coefficient to zero; previously selected peers are not revisited.

\FloatBarrier
\section{Experimental Setup and Reproducibility}
\label{app:experimental-setup}

\subsection{Models and Teacher--Student Configurations}
\label{app:models}

Section~\ref{sec:experiments:setup} introduces the student and domain
teachers. The specific Math and Code teacher checkpoints are
Qwen3-4B-Non-Thinking-RL-Math-Step500 and
Qwen3-4B-Non-Thinking-RL-Code-Step300, respectively
\citep{yang2026gopd}.
The medical teacher is II-Medical-8B, a reinforcement-learning (RL)
checkpoint based on Qwen3-8B \citep{intelligentinternet2025iimedical8b}.
The student undergoes full-parameter training; all teachers and the shared
reference remain frozen and are used only for scoring.

The models use compatible Qwen3 tokenizers and vocabularies, allowing the
scorers to evaluate identical response token IDs at identical prefixes.
Scoring checks require matching token IDs and log-probability tensor shapes.

\subsection{Training Data and Prompt Construction}
\label{app:training-data}

Each paired training set contains 10,000 prompts per domain, for 20,000
prompts in total, and is randomly shuffled before training. For Math, we
filter DeepMath \citep{he2025deepmath} to samples with difficulty level
$\geq 6$, yielding a pool of 57K candidates, and sample 10,000 training
prompts from this pool. For Code, we sample 10,000 prompts from
Eurus-RL-Code \citep{cui2025process}. The Medical pool contains 10,000
prompts from four sources, as detailed in Table~\ref{tab:app-training-data},
including 4,000 sampled from the MedMCQA training split
\citep{pal2022medmcqa}. The MedMCQA validation split is used only for
evaluation, not for constructing the training pool.

\begin{table}[!htbp]
    \caption{\textbf{Composition of the training prompt pools.} Counts denote
    sampled training prompts. Each paired setting uses the two corresponding
    domain pools.}
    \label{tab:app-training-data}
    \centering\small
    \begin{tabularx}{\textwidth}{@{}lXr@{}}
        \toprule
        Domain & Source & Sampled prompts \\
        \midrule
        Math & DeepMath \citep{he2025deepmath}, difficulty level $\geq 6$ (57K candidates) & 10,000 \\
        Code & Eurus-RL-Code \citep{cui2025process} & 10,000 \\
        \midrule
        Medical & MedMCQA training split \citep{pal2022medmcqa} & 4,000 \\
        & MedReason \citep{wu2025medreason} & 3,000 \\
        & Medical-R1-Distill-Data
          \citep{freedomintelligence2025medicalr1,chen2024huatuogpto1} & 2,000 \\
        & m23k \citep{huang2025m1} & 1,000 \\
        \bottomrule
    \end{tabularx}
\end{table}

Domain labels select the routed teacher; only the prompt is supplied as the
model input. Ground-truth answers are not used as outcome rewards in the
distillation loss. Medical prompts request step-by-step reasoning with the
final answer enclosed in \texttt{\textbackslash boxed\{\}}. Thinking mode is
disabled through the chat template, although generated responses can still
contain intermediate reasoning.

\subsection{Training Hyperparameters and Infrastructure}
\label{app:hyperparameters}

Table~\ref{tab:app-training-config} records the shared optimization and
rollout settings of the archived \method runs. The implementation uses
verl with an FSDP student and vLLM for
rollout generation and teacher scoring. Baseline-specific update
schedules are described in Appendix~\ref{app:baseline-details}.

\begin{table}[!htbp]
    \caption{\textbf{Shared optimization and rollout settings of \method.}
    Each rollout batch produces one optimizer update.}
    \label{tab:app-training-config}
    \centering\small
    \begin{tabularx}{\textwidth}{@{}lX@{}}
        \toprule
        Hyperparameter & Value \\
        \midrule
        Training / PPO minibatch size & 1,008 / 1,008 prompts \\
        Rollout batches (paired / three-domain) & 58 / 116 \\
        PPO epochs per rollout batch & 1 \\
        Responses per prompt & 1 \\
        Rollout temperature / top-\(p\) & 1.0 / 1.0 \\
        Maximum prompt / response length & 2,048 / 16,384 tokens \\
        Optimizer & AdamW \\
        Learning rate / schedule & \(10^{-5}\) / constant, no warm-up \\
        Adam betas / weight decay & \((0.9,0.999)\) / 0.01 \\
        Global gradient norm clipping & 1.0, after merging domain gradients \\
        Advantage clipping & \([-10,10]\), full method and all ablations \\
        PPO ratio-clipping parameter & 0.2 \\
        Distillation estimator & Sampled-token \(k_1\), policy-gradient surrogate \\
        Task reward / separate trainer KL reward & Disabled / disabled \\
        Checkpoint interval & 29 optimizer steps \\
        In-training evaluation & Disabled; offline checkpoint evaluation \\
        \bottomrule
    \end{tabularx}
\end{table}

The shared modulation parameters are \(\tau=1\),
\(\rho_{\mathrm g}=0.5\),
and \([\lambda_{\min},\lambda_{\max}]=[0.5,1.5]\), with
\(10^{-12}\) for numerical stability. Paired-domain training uses
bidirectional projection with \(m=0.6\). Because agreement lies in \([-1,1]\),
the unbounded mapping \(1+0.5s_t\) already lies in \([0.5,1.5]\);
clipping does not further restrict this mapping in exact arithmetic.

The paired- and three-domain configurations use one node with eight NVIDIA
B200 GPUs. Six GPUs host the FSDP student and three rollout replicas with
tensor-parallel size two. The remaining two GPUs host teacher scoring. In
paired settings, one 4B teacher shares a GPU with the frozen reference; in
three-domain training, the Math teacher, Code teacher, and reference share
one GPU, while the Medical teacher occupies the other. Gradient checkpointing,
padding removal, and dynamic batching are enabled. The three-domain
\method sampler forms balanced batches of 336 prompts per domain.

The audited ExOPD, Open-MOPD, and CaMOPD runs use 1,008 prompts per rollout
batch, with 58 batches in paired settings and 116 in three-domain settings.
ExOPD and CaMOPD perform one optimizer update per batch. Open-MOPD divides
that batch into four sequential minibatches of 252 prompts, giving 232 and
464 optimizer updates, respectively, while retaining one response per prompt.
Thus, matching rollout budgets does not imply matching optimizer-update
counts; CaMOPD also changes the allocation of prompts across domains through
its alternating schedule.
\FloatBarrier

\paragraph{Student sizes, reference models, and Single OPD.}
For both \method and ExOPD, the reference is a frozen copy of the student's
initialization: Qwen3-4B in 4B experiments and Qwen3-1.7B in 1.7B experiments.
For Qwen3-1.7B, Table~\ref{tab:main-results}(a)--(b) reports multi-domain
checkpoints at step 58; Student denotes the original,
undistilled model. Open-MOPD uses the sampled-token \(k_1\) estimator.
Within each student size, Math, Code, and Medical Single OPD share the same
training configuration and differ only in their domain-specific datasets.
All three Qwen3-1.7B Single OPD runs start from the original model and train
for 50 steps, evaluated with Avg@32, Avg@4, and Avg@1, respectively.
Qwen3-4B Single OPD uses the matching in-domain checkpoints in
Table~\ref{tab:single-domain-results}.

\subsection{Baseline Implementations}
\label{app:baseline-details}

We implement the baselines within the same MyOPD/verl training stack, using
the corresponding Qwen3-4B or Qwen3-1.7B initialization and domain teachers introduced in
Section~\ref{sec:experiments:setup}. Teacher scoring uses the sampled tokens
of student-generated responses. The baseline paths use the sampled-token
\(k_1\) reverse-KL estimator and a clipped policy-gradient surrogate with
clipping parameter 0.2; task rewards, entropy regularization, and a separate
trainer KL reward are disabled. The method-specific implementations are as
follows.

\paragraph{Vanilla MOPD.}
Each prompt is routed to its domain teacher
\citep{ma2026mopdmultiteacheronpolicydistillation}. The token advantage
\(\log\pi_{T_d}(y_t\mid h_t)-\log\pi_\theta(y_t\mid h_t)\) is clipped to
\([-10,10]\) and detached, without cross-teacher modulation or gradient
projection. The archived Math--Code and Code--Medical runs each use
58 optimizer updates, while the three-domain run uses 116.

\paragraph{ExOPD.}
We use a fixed extrapolation coefficient of 1.25.
The reference is the frozen student initialization, Qwen3-4B or Qwen3-1.7B
according to student size \citep{yang2026gopd}. For each sampled token,
the detached advantage is
\(1.25(\log\pi_{T_d}-\log\pi_{\mathrm{ref}})
-(\log\pi_\theta-\log\pi_{\mathrm{ref}})\), with all log probabilities
computed at the same response prefix.
The cross-teacher gate, gradient projection, and domain-balancing modules
are disabled for this baseline.

\paragraph{CaMOPD.}
The paired-domain implementation alternates homogeneous domain batches and
selects examples using the rollout-time gap
\(\Delta_t=\log\pi_{T_d}(y_t\mid h_t)-\log\pi_{\mathrm{old}}(y_t\mid h_t)\)
\citep{chen2026counteractionawaremultiteacheronpolicydistillation}.
Recovery examples are scored by the mean of \(|\Delta_t|\), while
preservation examples use the mean of \(\max(\Delta_t,0)\). Within each
branch, examples are ranked by this score and the smallest prefix covering
80\% of the total score mass is retained. For Medical--Math and Medical--Code,
Medical is the recovery branch: the update ratio is 3:1, and selected-example
advantage scales are 2.0 for Medical and 1.0 for the other domain. Math--Code
uses Math as recovery and Code as preservation, a 1:1 update ratio, and unit
scales for both branches. The three-domain adaptation treats domains as peer
branches, using the positive-gap score, the same 0.8 score-mass threshold,
and unit selected-example weights. We sample Medical, Math, and Code data
in a 3:1:1 ratio, prioritizing the weaker Medical domain in line with
CaMOPD's capability-recovery objective.

\paragraph{Open-MOPD.}
We adapt its balancing and reward-refresh mechanisms to the sampled-token
OPD path \citep{gao2026openmopd}. At
rollout time, a domain's weight is proportional to its uniform target share
divided by its observed response-token share, multiplied by its mean absolute
teacher--student gap relative to the mean gap across domains. This gap factor
uses exponent 1.0 and is clipped to \([0.05,20]\). The weights are normalized
to have token-weighted mean one and remain fixed within the rollout batch.
Dynamic allocation is therefore implemented through domain loss weights.
Each batch is split into four sequential optimizer minibatches. On each
minibatch, the OPD advantage is recomputed using the current student's
log probabilities and cached teacher scores, providing reward refresh.

\paragraph{PCGrad.}
For the two-domain comparison \citep{yu2020gradient}, separate backward passes
mask the shared minibatch to each domain's own responses. For each trainable
parameter tensor, the implementation computes the two gradients' dot product
and squared norms across FSDP shards. If the dot product is negative, both
gradients are projected against the original opposing gradient; the two
corrected gradients are then summed. This symmetric two-task calculation
requires no random task ordering. Gradient snapshots use BF16 and dot products
and squared norms use FP64. The comparison applies global gradient clipping
after merging, followed by one optimizer
step. These gradients are formed on their respective domains' responses,
following the independent-domain comparison in
Section~\ref{sec:experiments:analysis}; same-trajectory auxiliary gradients
are not used for this projection.
The archived PCGrad runs whose benchmark scores appear in
Table~\ref{tab:app-benchmark-details} also enable reward-agreement
modulation and are marked with a dagger in that table.

\subsection{Evaluation Protocols}
\label{app:evaluation-protocols}

We evaluate 11 benchmarks across Code (HumanEval+, MBPP+, and LiveCodeBench)
\citep{liu2023evalplus,jain2024livecodebench}, Math (AIME 2024/2025 and HMMT
February/November 2025)
\citep{maa2024aime,maa2025aime,hmmt2025feb,hmmt2025nov},
and Medical (MedQA, MedXpertQA, MedMCQA, and SuperGPQA-Med)
\citep{jin2020medqa,zuo2025medxpertqa,pal2022medmcqa,mapteam2025supergpqa}.

\paragraph{Generation setup.}
Offline evaluation uses vLLM with temperature 1.0, top-\(p\) 1.0, and a
maximum of 16,384 generated tokens per completion. Chat prompts disable
thinking mode via \texttt{enable\_thinking=False}.

\paragraph{Mathematics.}
Each of the four Math test sets contains 30 questions, with 32 sampled
completions per question. We append the instruction ``Please reason step
by step, and put your final answer within \texttt{\textbackslash boxed\{\}}.''
The evaluator extracts the last boxed answer and uses \texttt{math-verify}
to compare it with the reference answer; a missing or malformed boxed answer
is incorrect. The reported Avg@32 is the mean correctness over completions,
averaged over questions:
\begin{equation}
    \operatorname{Avg@32}
      =\frac{1}{N}\sum_{i=1}^{N}\frac{1}{32}
         \sum_{r=1}^{32}\mathbf{1}[\hat a_{i,r}=a_i].
    \label{eq:app-avg32}
\end{equation}
Here equality denotes equivalence under the answer verifier. Avg@32 equals
the pass@1 estimate computed from the same 32 samples. This differs
from pass@32, which counts a question as solved when at least one of its
32 completions is correct.

\paragraph{Code.}
All three Code benchmarks use four completions per problem. We report
Avg@4, the fraction of passing completions averaged over problems, which
equals the pass@1 estimate from the same four samples. HumanEval+ contains
164 problems and MBPP+ contains 378. A completion must pass both the original
and additional EvalPlus tests to count as correct. Prompts request a Python
solution in a fenced code block. Our local evaluation protocol (v3) prioritizes
Python or untagged Markdown code blocks, selects the candidate with the most
nonempty lines, and uses tree-sitter to retain the target entry point and its
dependencies. Generation continues until EOS or the token limit, without
stopping at the first closing code fence. EvalPlus batches up to 256
generation requests across problems and sets each test's execution limit to
\(\max(5\,\mathrm{s},6t_{\mathrm{ref}})\), where \(t_{\mathrm{ref}}\) is the
reference solution's runtime for that test.

\paragraph{LiveCodeBench.}
We use the v6 code-generation split: 175 problems with contest dates from
January 4 to April 6, 2025, giving 700 completions in total. The local loader
selects \texttt{test6.jsonl}, with no additional date filtering. The runner
uses the Qwen3-4B non-thinking prompt template and extracts the final fenced
code block. Each completion is checked against the problem's public and
private tests with a 60-second per-test timeout. Compilation failures,
runtime errors, timeouts, and failed tests count as incorrect completions.

\paragraph{Medical.}
The evaluation uses one completion per question and reports Avg@1
multiple-choice accuracy. It covers the four-option MedQA-USMLE test set
(1,273 questions), MedXpertQA Text test set (2,450; ten options), MedMCQA
validation set (4,183; four options), and the SuperGPQA subset selected by
\texttt{discipline=Medicine} (2,755; ten options). Each prompt includes the
question and labeled options, and requests step-by-step reasoning with the
final answer letter in \texttt{\textbackslash boxed\{\}}. The parser first
checks the last boxed letter, then falls back to an explicit answer phrase
such as ``Answer: B'' or a terminal ``(B)''. Letters are normalized to
uppercase and must belong to the question's available options. Exact match
to the reference letter determines correctness; unparseable answers are
incorrect.

\paragraph{Aggregation.}
All scores are reported as percentages. Each benchmark weights its questions
equally, and domain averages are arithmetic means over the benchmarks within
that domain, irrespective of their different numbers of questions.
The 11-task average in Table~\ref{tab:single-domain-results} weights all
11 benchmarks equally; it is distinct from averaging the three domain means,
because Code contains three benchmarks and the other domains contain four.

\paragraph{Teacher-disagreement and gradient-conflict probe.}
For the fixed-checkpoint probe in
Figure~\ref{fig:cross-teacher-advantage-modulation} (left), we compute
$S_-=\operatorname{mean}_t[\max(-s_t,0)]$ over all valid response tokens
pooled across each four-trajectory batch, using
Eq.~(\ref{eq:teacher-reward-agreement}) with $\tau=1$.
We partition batches into five within-domain quintiles, with 50 batches
per quintile. A raw conflict is a negative cosine between
the full-parameter routed- and auxiliary-teacher OPD gradients on the same
batch, without token modulation, projection, or optimizer updates.
This five-bin analysis is exploratory; shading connects pointwise 95\%
Wilson interval limits. The probe is separate from the token-visualization
samples in Appendix~\ref{app:extended-token-visualizations}.

\FloatBarrier
\section{Additional Experimental Results}
\label{app:additional-results}

\subsection{Detailed Benchmark Results}
\label{app:benchmark-details}

Table~\ref{tab:app-benchmark-details} provides available benchmark breakdowns
for the ablations and PCGrad comparison.

\begin{table}[!htbp]
    \caption{\textbf{Per-benchmark breakdown of the ablations and PCGrad comparison.}
    Avg. is the benchmark mean; rounding may cause small differences from the
    mean of displayed scores.
    Scores are percentages (higher is better).}
    \label{tab:app-benchmark-details}
    \centering\scriptsize
    \setlength{\tabcolsep}{2pt}
    \renewcommand{\arraystretch}{1.12}
    \begin{tabularx}{\textwidth}{@{}p{98pt}*{10}{>{\centering\arraybackslash}X}@{}}
        \toprule
        \rowcolor{resultsetting}
        \multicolumn{11}{l}{\textbf{(a) Math $\times$ Medical}} \\
        \midrule
        & \multicolumn{5}{c}{Math} & \multicolumn{5}{c}{Medical} \\
        \cmidrule(lr){2-6}\cmidrule(l){7-11}
        Method & AIME24 & AIME25 & \shortstack{HMMT\\25 Feb.}
        & \shortstack{HMMT\\25 Nov.} & Avg.
        & MedQA & \shortstack{Med\\Xpert\\QA} & \shortstack{Med-\\MCQA}
        & \shortstack{GPQA\\Med} & Avg. \\
        \midrule
        w/o Credit Assign. & 59.69 & 53.54 & 32.71 & 37.19 & 45.78 & 76.98 & 15.88 & 62.08 & 36.01 & 47.74 \\
        w/o Grad. Protection & 61.46 & 54.48 & 33.75 & 39.06 & 47.19 & 76.04 & 15.39 & 63.28 & 37.71 & 48.11 \\
        w/o Adaptive Retention & 58.13 & 55.42 & 33.02 & 36.56 & 45.78 & 77.53 & 15.59 & 63.61 & 36.84 & 48.40 \\
        PCGrad$^\dagger$ & 62.60 & 53.54 & 33.75 & 37.50 & 46.85 & 77.93 & 14.65 & 62.54 & 36.99 & 48.03 \\
        \midrule
        \rowcolor{resultavg}
        \method & 63.85 & 53.96 & 33.65 & 38.96 & 47.60 & 77.06 & 15.67 & 62.83 & 37.35 & 48.23 \\
        \bottomrule
    \end{tabularx}
    \par\medskip
    \begin{tabularx}{\textwidth}{@{}p{98pt}*{3}{>{\centering\arraybackslash}X}>{\centering\arraybackslash}p{43pt}*{4}{>{\centering\arraybackslash}X}>{\centering\arraybackslash}p{43pt}@{}}
        \toprule
        \rowcolor{resultsetting}
        \multicolumn{10}{l}{\textbf{(b) Code $\times$ Medical}} \\
        \midrule
        & \multicolumn{4}{c}{Code} & \multicolumn{5}{c}{Medical} \\
        \cmidrule(lr){2-5}\cmidrule(l){6-10}
        Method & \shortstack{Human\\Eval+} & MBPP+ & LCB & Avg.
        & MedQA & \shortstack{Med\\Xpert\\QA} & \shortstack{Med-\\MCQA}
        & \shortstack{GPQA\\Med} & Avg. \\
        \midrule
        w/o Credit Assign. & 84.15 & 69.05 & 25.43 & 59.54 & 76.20 & 16.08 & 61.77 & 35.79 & 47.46 \\
        w/o Grad. Protection & 87.65 & 71.96 & 29.43 & 63.01 & 75.18 & 14.16 & 61.53 & 36.62 & 46.87 \\
        w/o Adaptive Retention & 86.74 & 73.68 & 28.14 & 62.85 & 76.12 & 14.08 & 62.25 & 35.68 & 47.03 \\
        PCGrad$^\dagger$ & 84.45 & 70.11 & 28.14 & 60.90 & 76.90 & 15.71 & 62.49 & 35.17 & 47.57 \\
        \midrule
        \rowcolor{resultavg}
        \method & 87.65 & 72.69 & 28.71 & 63.02 & 75.73 & 15.96 & 62.97 & 36.33 & 47.75 \\
        \bottomrule
    \end{tabularx}
    \par\smallskip
    {\raggedright $\dagger$ The archived PCGrad runs also enable
    reward-agreement modulation.\par}
\end{table}

\FloatBarrier

\paragraph{Single-domain reference results.}
Table~\ref{tab:single-domain-results} lists the available single-domain
checkpoints across all 11 benchmarks, complementing
Figure~\ref{fig:gradient-conflict}.
\begin{table}[!htbp]
    \caption{\textbf{Cross-domain evaluation of single-domain training checkpoints.}
    Each Qwen3-4B single-domain model is trained with OPD on 10,000 in-domain prompts for
    3 epochs and evaluated across all 11 benchmarks.
    The best and second-best results among the three checkpoints are shown in bold
    and underlined, respectively.}
    \label{tab:single-domain-results}
    \centering
    \footnotesize
    \setlength{\tabcolsep}{3.2pt}
    \setlength{\aboverulesep}{0pt}
    \setlength{\belowrulesep}{0pt}
    \renewcommand{\arraystretch}{0.96}
    \definecolor{singledomainsetting}{RGB}{232,239,243}
    \definecolor{singledomainavg}{RGB}{244,244,244}
    \begin{tabularx}{\textwidth}{@{}l*{5}{>{\centering\arraybackslash}X}@{}}
        \toprule
        \textbf{Benchmark}
        & \textbf{Teacher}
        & \textbf{Student}
        & \textbf{Math-only}
        & \textbf{Medical-only}
        & \textbf{Code-only} \\
        \midrule
        \rowcolor{singledomainsetting}
        \multicolumn{6}{l}{\textbf{Code}} \\
        \rowcolor{white}
        HumanEval+       & 85.98 & 76.68 & \secondscore{81.40} & 81.10 & \bestscore{84.30} \\
        MBPP+            & 79.63 & 65.08 & 69.71 & \bestscore{72.95} & \secondscore{70.44} \\
        LCB v6           & 27.00 & 17.14 & 22.00 & \secondscore{22.14} & \bestscore{28.00} \\
        \rowcolor{singledomainavg}
        \textit{Avg.}    & 64.20 & 52.97 & 57.70 & \secondscore{58.73} & \bestscore{60.91} \\
        \midrule
        \rowcolor{singledomainsetting}
        \multicolumn{6}{l}{\textbf{Math}} \\
        \rowcolor{white}
        AIME24           & 57.71 & 21.46 & \secondscore{58.13} & \bestscore{59.38} & 43.75 \\
        AIME25           & 55.00 & 21.77 & \bestscore{51.67} & \secondscore{46.15} & 37.29 \\
        HMMT25 Feb.      & 32.29 & 10.00 & \bestscore{31.56} & \secondscore{30.31} & 20.00 \\
        HMMT25 Nov.      & 39.27 & 8.23 & \secondscore{35.73} & \bestscore{36.56} & 24.79 \\
        \rowcolor{singledomainavg}
        \textit{Avg.}    & 46.07 & 15.37 & \bestscore{44.27} & \secondscore{43.10} & 31.46 \\
        \midrule
        \rowcolor{singledomainsetting}
        \multicolumn{6}{l}{\textbf{Medical}} \\
        \rowcolor{white}
        MedQA            & 87.90 & 64.73 & \secondscore{70.86} & \bestscore{77.06} & 67.71 \\
        MedXpertQA       & 25.71 & 13.80 & \secondscore{14.90} & \bestscore{15.31} & 14.00 \\
        MedMCQA          & 71.38 & 55.63 & \secondscore{60.17} & \bestscore{62.44} & 57.33 \\
        SuperGPQA-Med    & 47.19 & 29.87 & \secondscore{34.70} & \bestscore{36.81} & 33.32 \\
        \rowcolor{singledomainavg}
        \textit{Avg.}    & 58.05 & 41.01 & \secondscore{45.16} & \bestscore{47.90} & 43.09 \\
        \midrule
        \rowcolor{singledomainavg}
        \textbf{11-Task Avg.}
        & 55.37
        & 34.94
        & \secondscore{48.26}
        & \bestscore{49.11}
        & 43.72 \\
        \bottomrule
    \end{tabularx}
\end{table}

\FloatBarrier

\subsection{Full Results for Low-Conflict and Three-Domain Settings}
\label{app:low-conflict-results}
\label{app:three-domain-results}

Tables~\ref{tab:app-low-conflict-results} and~\ref{tab:app-three-domain-results}
provide the full per-benchmark Qwen3-4B results underlying the summaries in
Tables~\ref{tab:low-conflict-results} and~\ref{tab:three-domain-results},
respectively. Domain averages weight benchmarks equally. Single OPD uses
the matching in-domain checkpoints from Table~\ref{tab:single-domain-results}.

\begin{table}[!htbp]
    \caption{\textbf{Full results for low-conflict Math $\times$ Code.}
    Per-benchmark Qwen3-4B scores underlying Table~\ref{tab:low-conflict-results}.
    Scores are percentages (higher is better); Avg. weights benchmarks equally within each domain.
    Bold and underlining mark the best and second-best multi-domain methods, respectively.}
    \label{tab:app-low-conflict-results}
    \centering\footnotesize
    \setlength{\tabcolsep}{3.2pt}
    \renewcommand{\arraystretch}{1.05}
    \begin{tabularx}{\textwidth}{@{}l*{7}{>{\centering\arraybackslash}X}@{}}
        \toprule
        Benchmark & Student & \shortstack{Single\\OPD} & \shortstack{Vanilla\\MOPD}
        & ExOPD & \shortstack{Open-\\MOPD} & CaMOPD & Ours \\
        \midrule
        \rowcolor{resultsetting}
        \multicolumn{8}{l}{\textbf{Math}} \\
        AIME24 & 21.46 & 58.13 & 60.21 & \bestscore{63.33} & 56.77 & \secondscore{61.04} & 60.73 \\
        AIME25 & 21.77 & 51.67 & 55.21 & \bestscore{57.71} & 53.33 & 54.17 & \secondscore{57.40} \\
        HMMT25 Feb. & 10.00 & 31.56 & 31.98 & \bestscore{33.85} & 30.52 & 31.87 & \secondscore{33.33} \\
        HMMT25 Nov. & 8.23 & 35.73 & 37.08 & \secondscore{38.23} & 36.67 & 38.12 & \bestscore{39.48} \\
        \rowcolor{resultavg}
        Avg. & 15.37 & 44.27 & 46.12 & \bestscore{48.28} & 44.32 & 46.30 & \secondscore{47.73} \\
        \midrule
        \rowcolor{resultsetting}
        \multicolumn{8}{l}{\textbf{Code}} \\
        HumanEval+ & 76.68 & 84.30 & 85.06 & \secondscore{86.59} & 83.08 & 84.91 & \bestscore{86.89} \\
        MBPP+ & 65.08 & 70.44 & 70.37 & \secondscore{71.16} & 70.37 & 70.50 & \bestscore{72.75} \\
        LCB v6 & 17.14 & 28.00 & \secondscore{28.29} & 27.71 & 26.00 & 25.71 & \bestscore{29.57} \\
        \rowcolor{resultavg}
        Avg. & 52.97 & 60.91 & 61.24 & \secondscore{61.82} & 59.82 & 60.37 & \bestscore{63.07} \\
        \bottomrule
    \end{tabularx}
\end{table}

\providecommand{\resultblank}{\phantom{00.00}}
\providecommand{\bestscore}[1]{\textbf{#1}}
\providecommand{\secondscore}[1]{\underline{#1}}
\definecolor{threedomainsetting}{RGB}{232,239,243}
\definecolor{threedomainavg}{RGB}{244,244,244}

\begin{table}[!htbp]
    \caption{\textbf{Full results for generalization to three domains.}
    Per-benchmark Qwen3-4B scores underlying Table~\ref{tab:three-domain-results}.
    Scores are percentages (higher is better); Avg. weights benchmarks equally within each domain.
    Single OPD uses the matching in-domain checkpoint from Table~\ref{tab:single-domain-results}.
    The best and second-best results among the compared multi-domain methods are shown in bold and underlined, respectively.}
    \label{tab:app-three-domain-results}
    \centering
    \footnotesize
    \setlength{\tabcolsep}{3.2pt}
    \setlength{\dashlinedash}{1.5pt}
    \setlength{\dashlinegap}{1.5pt}
    \setlength{\aboverulesep}{0pt}
    \setlength{\belowrulesep}{0pt}
    \renewcommand{\arraystretch}{0.96}
    \begin{tabularx}{\textwidth}{@{}l*{2}{>{\centering\arraybackslash}X}*{5}{>{\centering\arraybackslash}X}@{}}
        \toprule
        \textbf{Benchmark}
        & \textbf{Student}
        & \shortstack{\textbf{Single}\\\textbf{OPD}}
        & \shortstack{\textbf{Vanilla}\\\textbf{MOPD}}
        & \textbf{ExOPD}
        & \shortstack{\textbf{Open-}\\\textbf{MOPD}}
        & \textbf{CaMOPD}
        & \textbf{Ours} \\
        \midrule
        \rowcolor{threedomainsetting}
        \multicolumn{8}{l}{\textbf{Code}} \\
        \rowcolor{white}
        HumanEval+       & 76.68 & 84.30 & \secondscore{86.28} & 84.76 & 85.52 & 84.45 & \bestscore{86.59} \\
        MBPP+            & 65.08 & 70.44 & 70.63 & \secondscore{71.23} & 70.44 & 70.90 & \bestscore{72.95} \\
        LCB              & 17.14 & 28.00 & 26.86 & \secondscore{28.57} & 25.71 & 27.57 & \bestscore{29.29} \\
        \rowcolor{threedomainavg}
        \textit{Avg.}    & 52.97 & 60.91 & 61.26 & \secondscore{61.52} & 60.56 & 60.97 & \bestscore{62.94} \\
        \midrule
        \rowcolor{threedomainsetting}
        \multicolumn{8}{l}{\textbf{Math}} \\
        \rowcolor{white}
        AIME24           & 21.46 & 58.13 & \secondscore{61.04} & 59.90 & 55.73 & 59.48 & \bestscore{61.88} \\
        AIME25           & 21.77 & 51.67 & \bestscore{57.50} & \secondscore{55.31} & 54.17 & 54.90 & 54.48 \\
        HMMT25 Feb.      & 10.00 & 31.56 & 31.67 & \bestscore{34.90} & 31.25 & 31.98 & \secondscore{33.54} \\
        HMMT25 Nov.      & 8.23 & 35.73  & 37.71 & \secondscore{38.23} & 35.10 & 37.50 & \bestscore{39.27} \\
        \rowcolor{threedomainavg}
        \textit{Avg.}    & 15.37 & 44.27 & 46.98 & \secondscore{47.09} & 44.06 & 45.96 & \bestscore{47.29} \\
        \midrule
        \rowcolor{threedomainsetting}
        \multicolumn{8}{l}{\textbf{Medical}} \\
        \rowcolor{white}
        MedQA            & 64.73 & 77.06 & 64.49 & 73.29 & 60.17 & \secondscore{76.75} & \bestscore{77.61} \\
        MedXpertQA       & 13.80 & 15.31 & 12.78 & 14.57 & 12.78 & \bestscore{16.69} & \secondscore{15.96} \\
        MedMCQA          & 55.63 & 62.44 & 57.14 & 60.94 & 55.06 & \secondscore{62.54} & \bestscore{63.11} \\
        SuperGPQA-Med    & 29.87 & 36.81 & 29.69 & 33.68 & 27.30 & \secondscore{36.81} & \bestscore{37.10} \\
        \rowcolor{threedomainavg}
        \textit{Avg.}    & 41.01 & 47.90 & 41.03 & 45.62 & 38.83 & \secondscore{48.20} & \bestscore{48.45} \\
        \bottomrule
    \end{tabularx}
\end{table}

\FloatBarrier

\subsection{Extended Token-Level Visualizations}
\label{app:extended-token-visualizations}

\begin{figure}[!htbp]
    \centering
    \includegraphics[width=\linewidth]{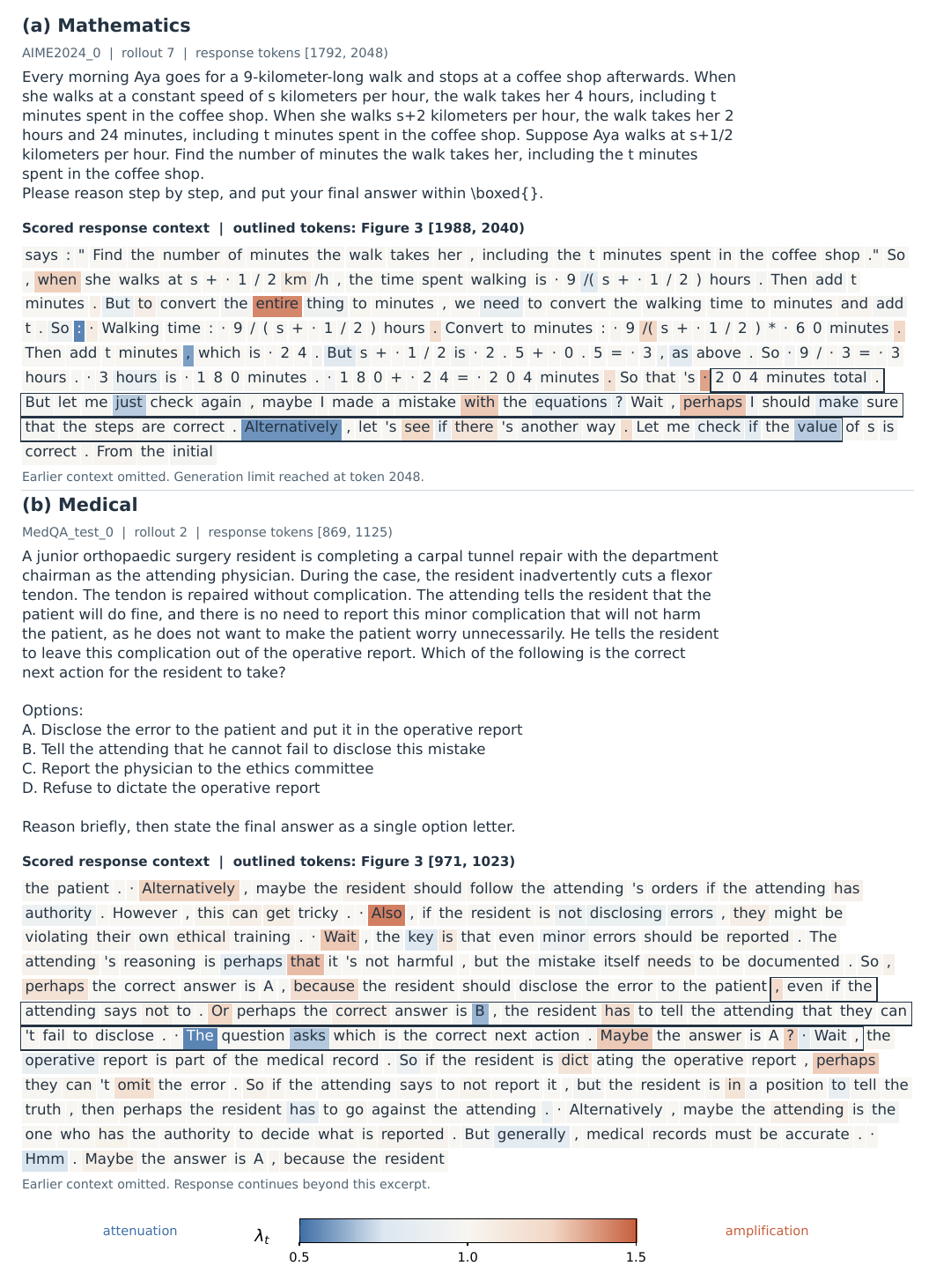}
    \caption{\textbf{Extended Math and Medical examples from the right panels of
    Figure~\ref{fig:cross-teacher-advantage-modulation}.}
    Outlines identify the main-text windows within the longer contexts.
    Blue and orange denote attenuation and amplification on the unchanged
    scale \(\lambda_t\in[0.5,1.5]\). Token indices are zero-based, with
    half-open intervals. Omitted context and the generation limit are marked.}
    \label{fig:app-extended-token-examples}
\end{figure}

Figure~\ref{fig:app-extended-token-examples} expands the exact trajectories
shown in the right panels of Figure~\ref{fig:cross-teacher-advantage-modulation} to 256-token
contexts. It includes the original prompts and marks the 52-token main-text
windows, preserving every displayed token's original coefficient. These
illustrative windows were selected from 20 sampled responses per prompt to
show both attenuation and amplification near an answer mention; they do not
estimate the overall coefficient distribution. The visualization samples
use the token-modulation-only checkpoint at step 58, temperature/top-\(p\)
\(1/1\), and a 2,048-token generation cap. Both selected
trajectories reached that cap.

\FloatBarrier

\begin{figure}[!htbp]
    \centering
    \includegraphics[width=\linewidth]{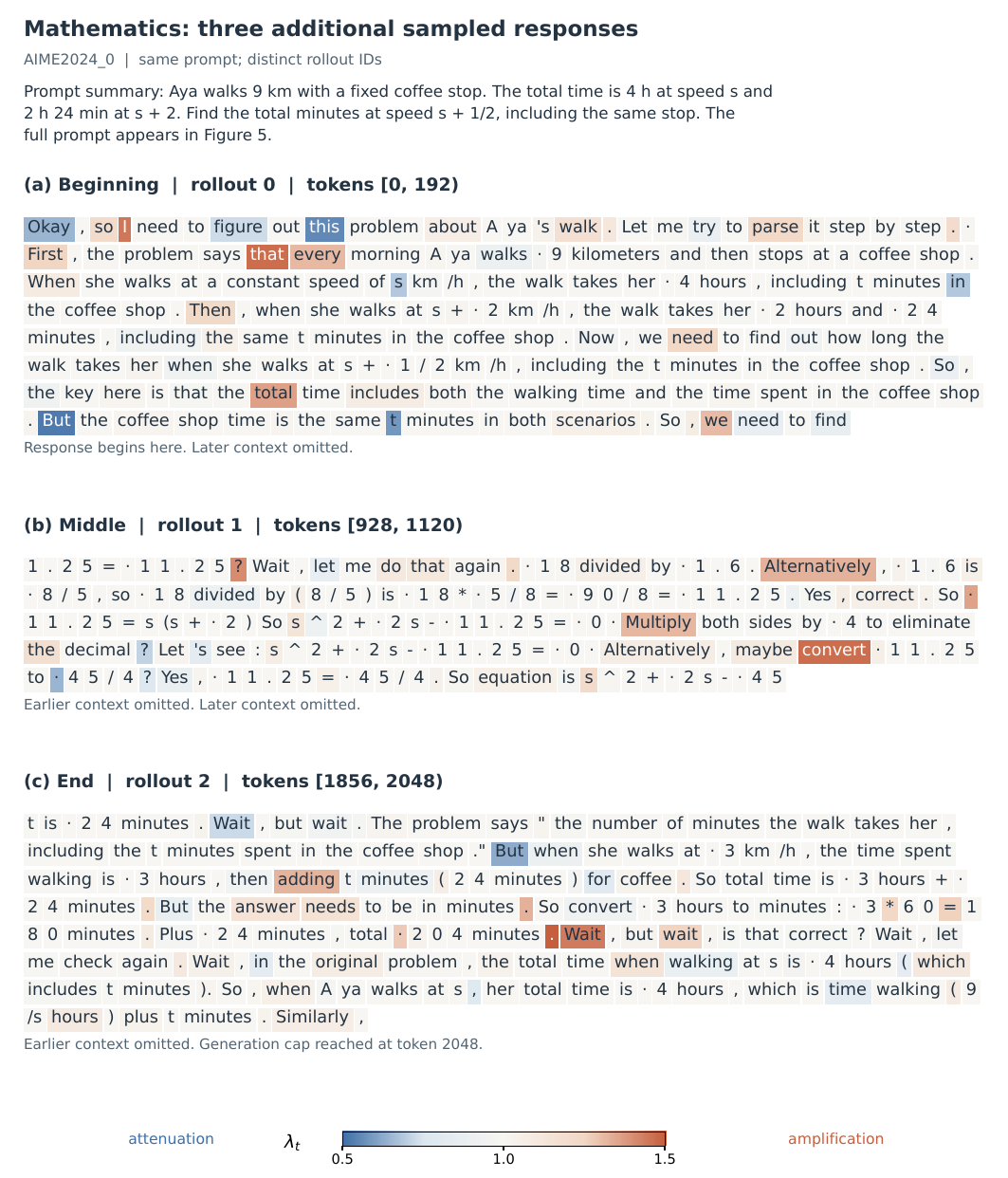}
    \caption{\textbf{Three additional Math response excerpts.}
    Panels show the beginning, middle, and end of three distinct sampled
    responses to \texttt{AIME2024\_0}. Original token coefficients and the
    \([0.5,1.5]\) color scale are preserved. A dot represents a whitespace-only
    token; omitted context and the generation cap are marked.}
    \label{fig:app-token-samples-math}
\end{figure}

\begin{figure}[!htbp]
    \centering
    \includegraphics[width=\linewidth]{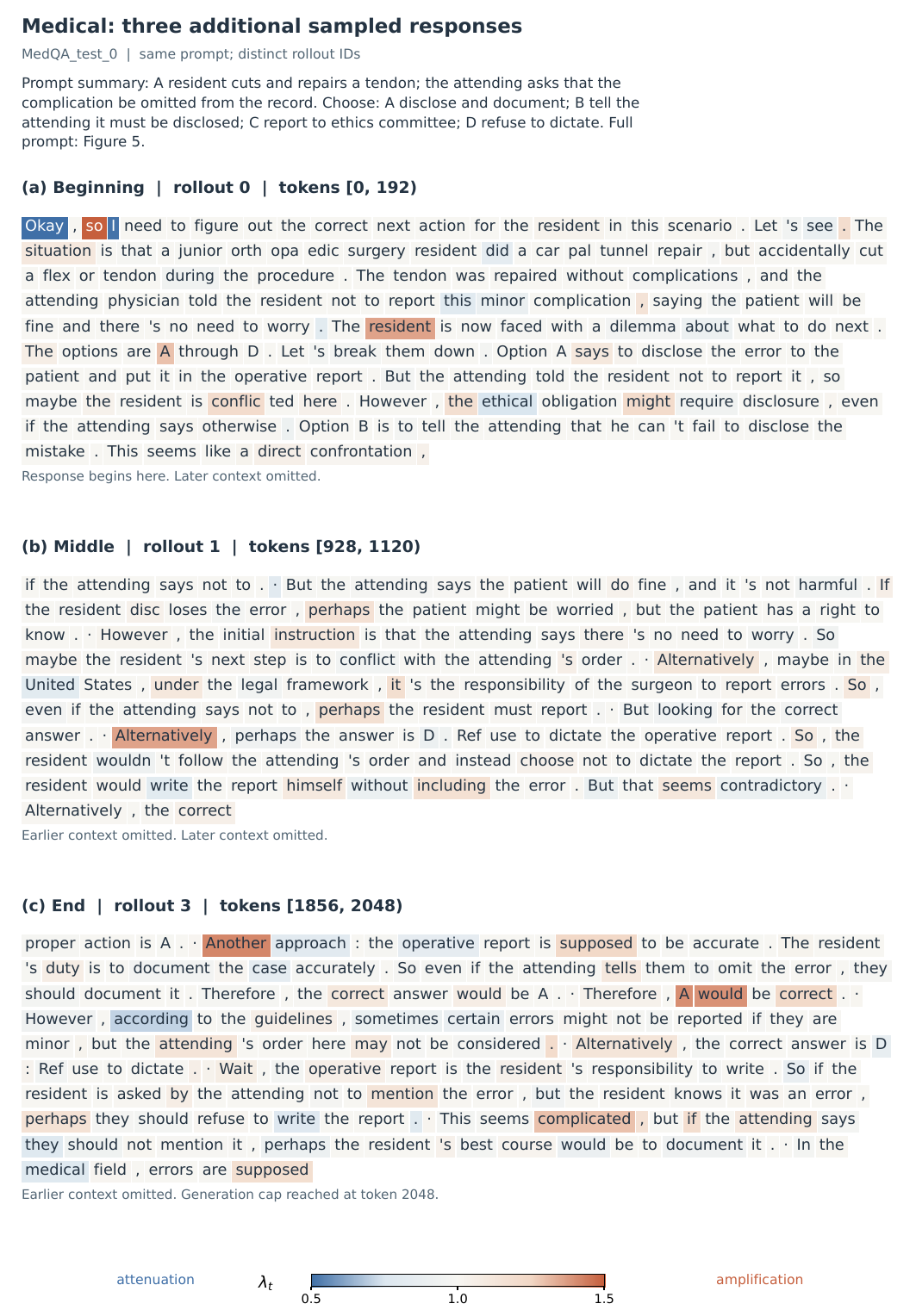}
    \caption{\textbf{Three additional Medical response excerpts.}
    Panels show the beginning, middle, and end of three distinct sampled
    responses to \texttt{MedQA\_test\_0}, using the same fixed-window rule
    and unchanged color scale as Figure~\ref{fig:app-token-samples-math}.
    These are generated reasoning excerpts, not verified answer explanations.}
    \label{fig:app-token-samples-medical}
\end{figure}

\paragraph{Additional fixed-window examples.}
Figures~\ref{fig:app-token-samples-math} and
\ref{fig:app-token-samples-medical} show three further responses per prompt.
We take the first three rollout IDs after excluding the response already
shown in Figure~\ref{fig:app-extended-token-examples}: Math IDs 0, 1, and 2;
Medical IDs 0, 1, and 3. In that order, we display the fixed 192-token
windows \([0,192)\), \([928,1120)\), and \([1856,2048)\).
This selection does not inspect coefficients or answer correctness.
Prompt summaries accompany the plots; the full prompts appear in
Figure~\ref{fig:app-extended-token-examples}.

\FloatBarrier

\FloatBarrier


\clearpage
\subsection{Training Dynamics}
\label{app:training-dynamics}

We complement the endpoint evaluations with complete training traces from
three archived \method runs: Math $\times$ Medical,
Code $\times$ Medical, and Math $\times$ Code.
Each run has 58 updates with 1,008 prompts per update, one PPO epoch,
and a 16,384-token response cap.
Both domains use reward-agreement modulation and tensor-wise adaptive
projection with norm-retention floor \(m=0.6\).
All plots retain every logged step without smoothing.

\begin{figure}[!htbp]
    \centering
    \includegraphics[width=\linewidth]{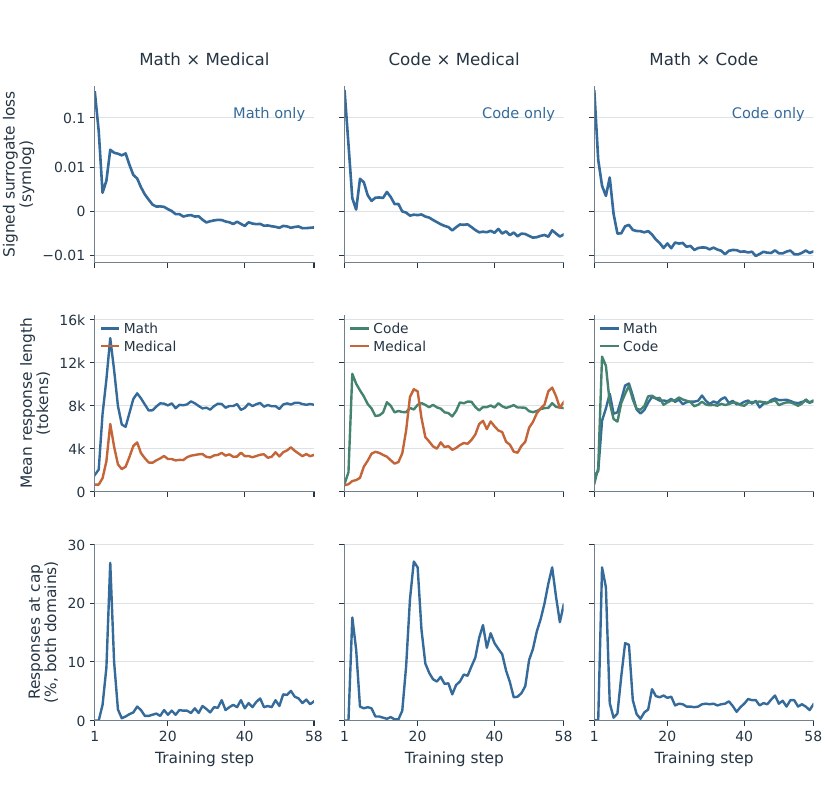}
    \caption{\textbf{Training loss and generation behavior across three
    domain pairings.} Columns show individual \method runs.
    Top: the logged task-0 surrogate contribution on a symmetric-log scale
    (linear within \(\pm0.01\)); each panel identifies the recorded domain.
    Middle: mean response length for each domain.
    Bottom: percentage of responses across both domains that reach
    16,384 tokens. Curves connect all 58 logged updates; negative loss
    values and fluctuations are preserved.}
    \label{fig:app-training-dynamics}
\end{figure}

\paragraph{Loss and generation behavior.}
Figure~\ref{fig:app-training-dynamics} shows the logged signed policy
surrogate, per-domain response lengths, and the fraction of responses
reaching the cap. The logger captures only the task-0 primary loss
contribution: Math in Math $\times$ Medical, and Code in the other two
runs. This quantity can be negative and is neither the full two-domain
objective nor a nonnegative KL estimate. It decreases sharply early in
training, while generation lengths and cap rates continue to vary.
These single-run traces describe training behavior; they do not establish
convergence of both domain objectives or comparative test accuracy.

\FloatBarrier

\begin{figure}[!htbp]
    \centering
    \includegraphics[width=\linewidth]{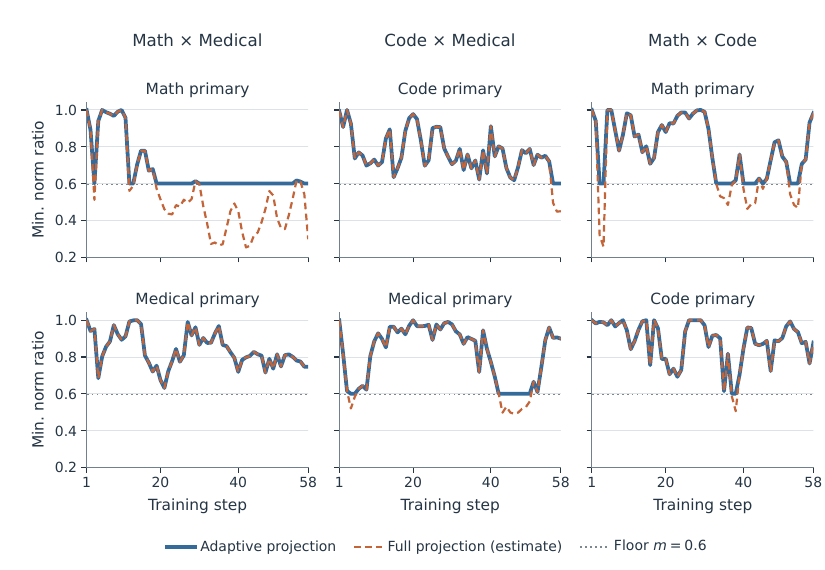}
    \caption{\textbf{Minimum norm retention over training.}
    Each panel identifies the primary domain whose gradient is projected;
    columns use the same runs as Figure~\ref{fig:app-training-dynamics}.
    Blue shows the logged ratio under adaptive projection; dashed orange
    shows the analytical full-projection estimate on the same gradients.
    The dotted line marks \(m=0.6\). Ratios are minima over conflicting
    parameter-tensor units, before merged-gradient clipping.
    Overlapping curves indicate identical minimum retention.
    All logged steps are shown without smoothing.}
    \label{fig:app-gradient-retention}
\end{figure}

\clearpage
\paragraph{Gradient-norm retention during projection.}
Figure~\ref{fig:app-gradient-retention} examines when the norm constraint
changes the projection strength. For each primary domain, the logger
computes two norm ratios from the same pre-projection gradients:
the ratio under full projection and the ratio under the chosen adaptive
strength. Each curve reports the minimum over conflicting parameter-tensor
units at that update (or one if none conflict). These are analytical
projection diagnostics before merging and global gradient clipping,
rather than measurements of the final optimizer displacement.

The adaptive ratios stay at or above \(0.6\), within numerical precision,
in all six domain traces. For example, in the Math branch of
Math $\times$ Medical, the full-projection estimate falls below \(0.6\)
in 40 of 58 updates and reaches a minimum of \(0.253\), whereas the
adaptive minimum is \(0.600\). The Medical branch of that same run
never crosses the floor, so the two curves coincide. This illustrates
how the constraint limits norm loss when needed without changing full
projection when its retained norm already satisfies the floor.
The full-projection curve uses the gradients encountered by \method;
it is not a separately trained baseline.

\FloatBarrier

\clearpage
\section{Limitations}
\label{app:limitations}

\paragraph{Computational overhead.}
\method incurs additional training overhead compared with Vanilla MOPD
because of auxiliary-teacher scoring and gradient processing.
Reusing student trajectories avoids additional rollout generation.
Future work will explore more efficient teacher evaluation and gradient
correction to reduce this overhead.

\paragraph{Limited evaluation of domain scalability.}
Our experiments cover mathematics, code, and medicine, with joint training
evaluated on at most three domains
(Table~\ref{tab:three-domain-results}). Future work will assess whether
the observed gains and training stability extend to broader domain
collections and larger teacher pools.


\end{document}